\documentclass[runningheads]{llncs}
\usepackage{graphicx}
\usepackage{tabularx, booktabs}
\usepackage{hyperref}
\newcommand{\etal} {\textit{et~al.}}
\usepackage[misc,geometry]{ifsym}

\begin{document}
\title{Unsupervised learning of text line segmentation by differentiating coarse patterns}
\titlerunning{Unsupervised learning of text line segmentation}
\author{Berat Kurar Barakat\inst{1}(\Letter)\and
Ahmad Droby\inst{1}\and
Raid Saabni \inst{2}\and
Jihad El-Sana\inst{1}}
\authorrunning{B. Kurar et al.}
\institute{Ben-Gurion University of the Negev \\
           Academic College of Telaviv Yafo \\
\email{\{berat, drobya, el-sana\}@post.bgu.ac.il},  
\email{raidsa@mta.ac.il}}


\maketitle 

\begin{abstract}
Despite recent advances in the field of supervised deep learning for text line segmentation, unsupervised deep learning solutions are beginning to gain popularity. In this paper, we present an unsupervised deep learning method that embeds document image patches to a compact Euclidean space where distances correspond to a coarse text line pattern similarity. Once this space has been produced, text line segmentation can be easily implemented using standard techniques with the embedded feature vectors. To train the model, we extract random pairs of document image patches with the assumption that neighbour patches contain a similar coarse trend of text lines, whereas if one of them is rotated, they contain different coarse trends of text lines. Doing well on this task requires the model to learn to recognize the text lines and their salient parts. The benefit of our approach is zero manual labelling effort. We evaluate the method qualitatively and quantitatively on several variants of text line segmentation datasets to demonstrate its effectivity.

\end{abstract}

\begin{figure*}[h]
\centering
\begin{tabular}
{m{0.13\textwidth} m{0.13\textwidth} m{0.13\textwidth} m{0.13\textwidth} m{0.13\textwidth}m{0.13\textwidth} m{0.13\textwidth}}
\includegraphics[width=0.13\textwidth, height=0.17\textwidth]{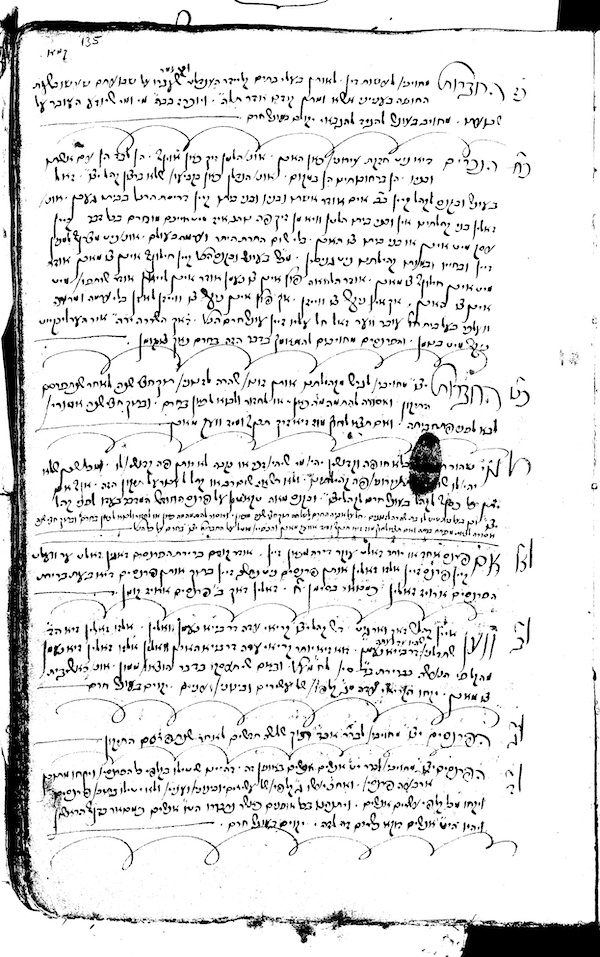} & 
\includegraphics[width=0.13\textwidth, height=0.17\textwidth]{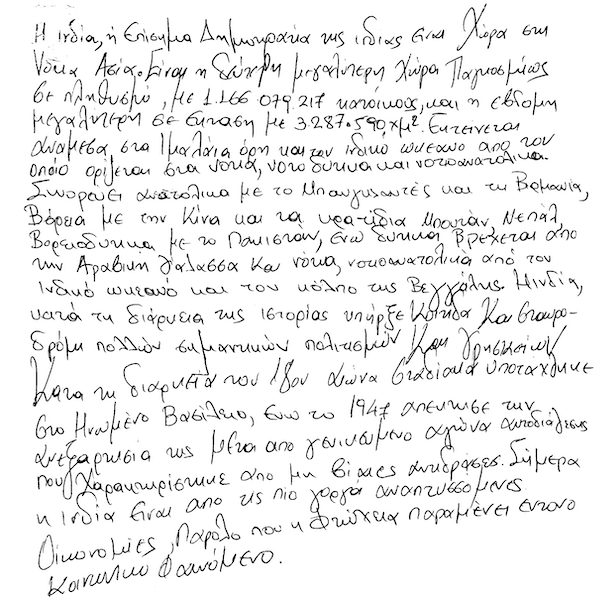} & 
\includegraphics[width=0.13\textwidth, height=0.17\textwidth]{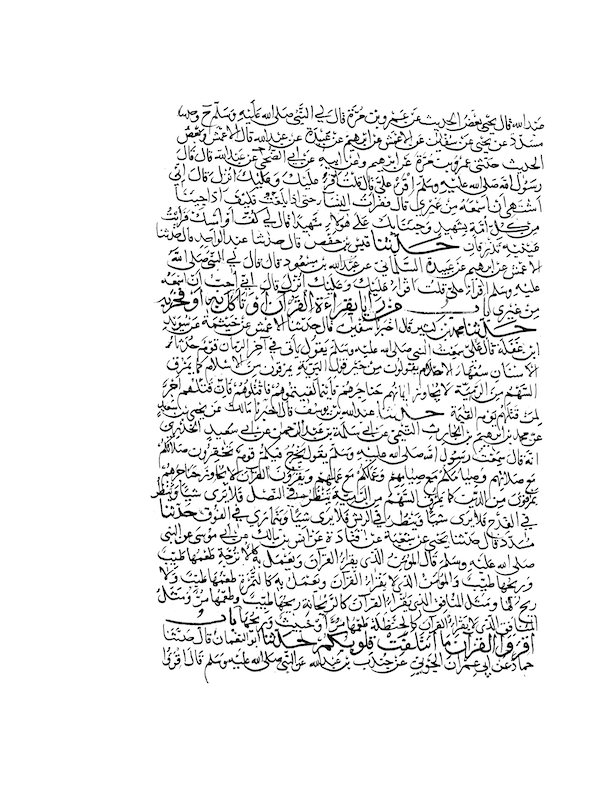} &
\includegraphics[width=0.13\textwidth, height=0.17\textwidth]{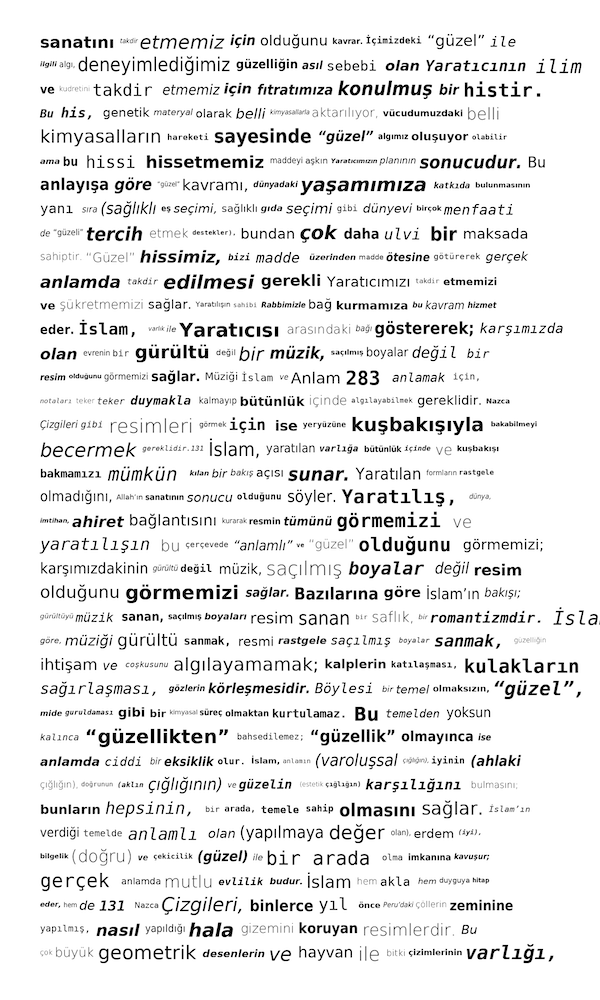} &
\includegraphics[width=0.13\textwidth, height=0.17\textwidth]{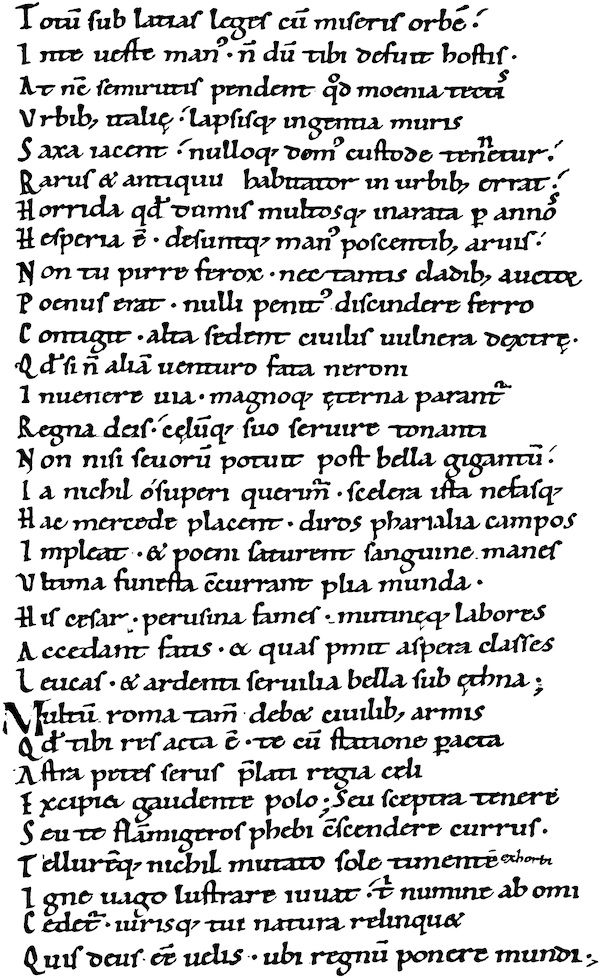} & 
\includegraphics[width=0.13\textwidth, height=0.17\textwidth]{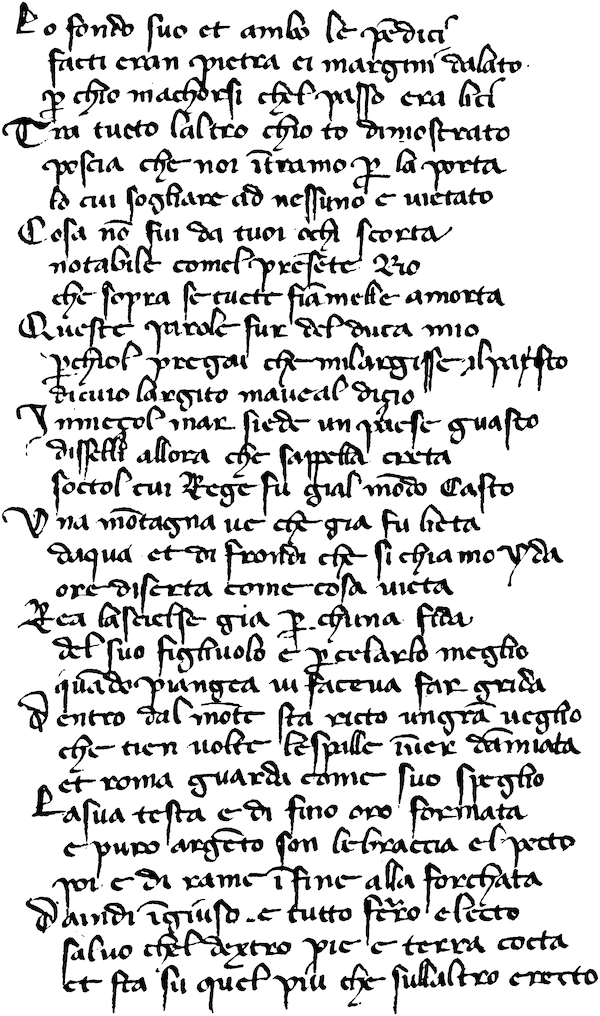} &
\includegraphics[width=0.13\textwidth, height=0.17\textwidth]{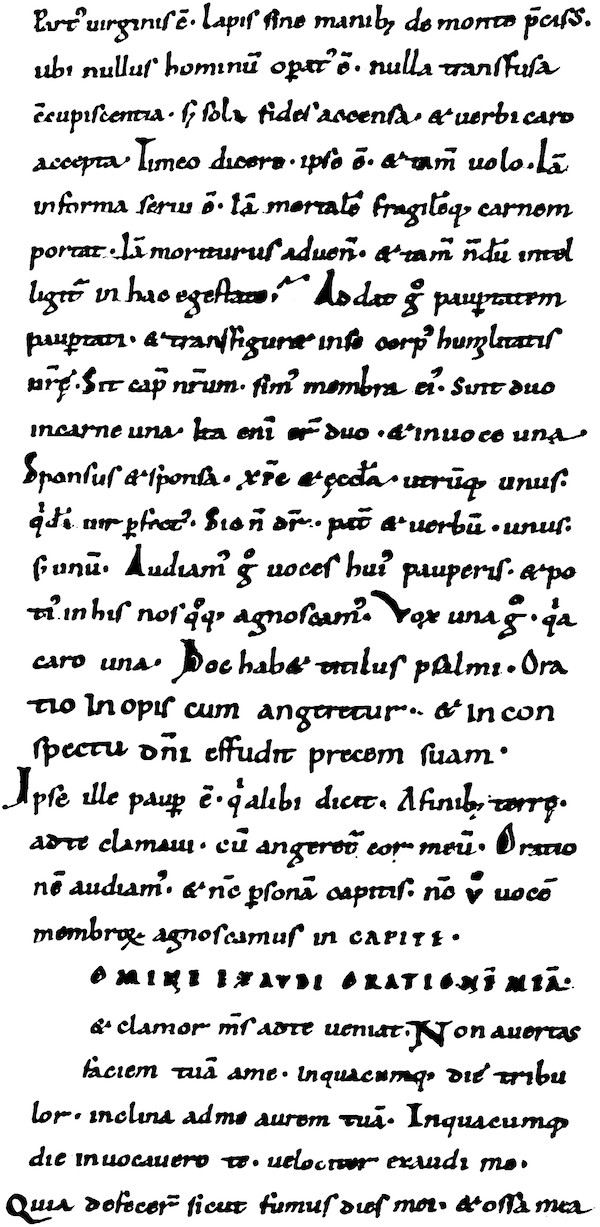} \\
\includegraphics[width=0.13\textwidth, height=0.17\textwidth]{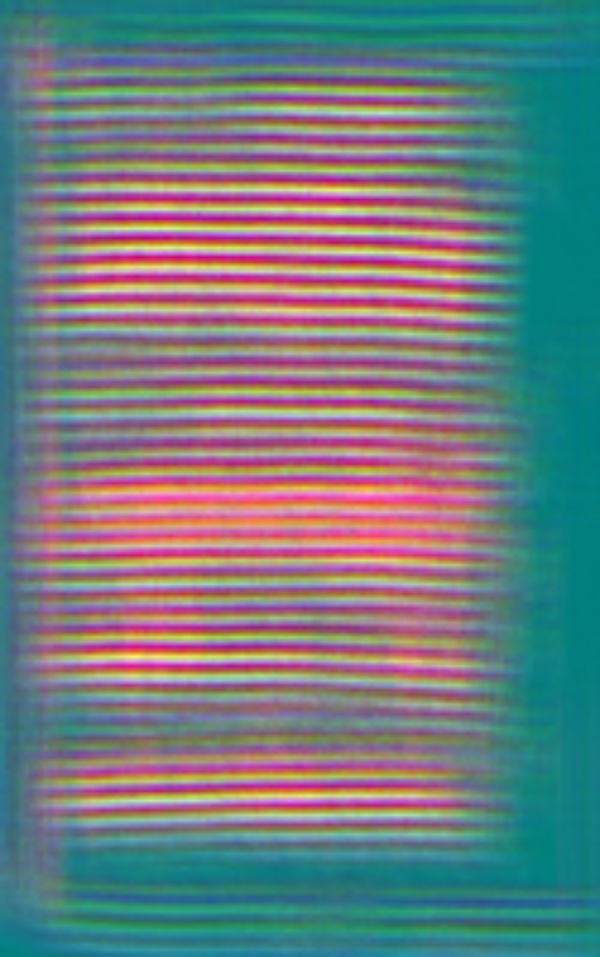} &
\includegraphics[width=0.13\textwidth, height=0.17\textwidth]{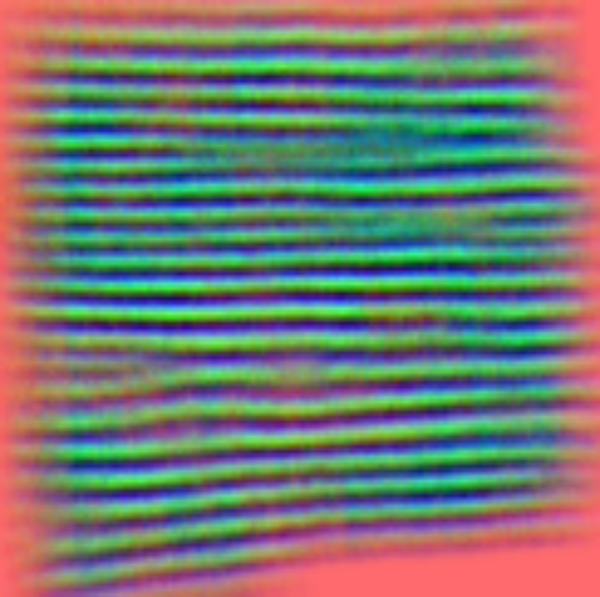} & 
\includegraphics[width=0.13\textwidth, height=0.17\textwidth]{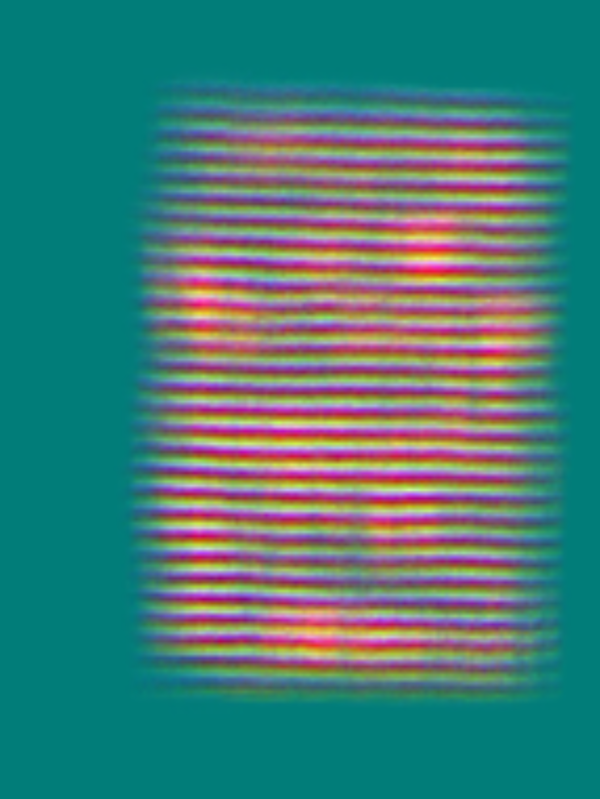} &
\includegraphics[width=0.13\textwidth, height=0.17\textwidth]{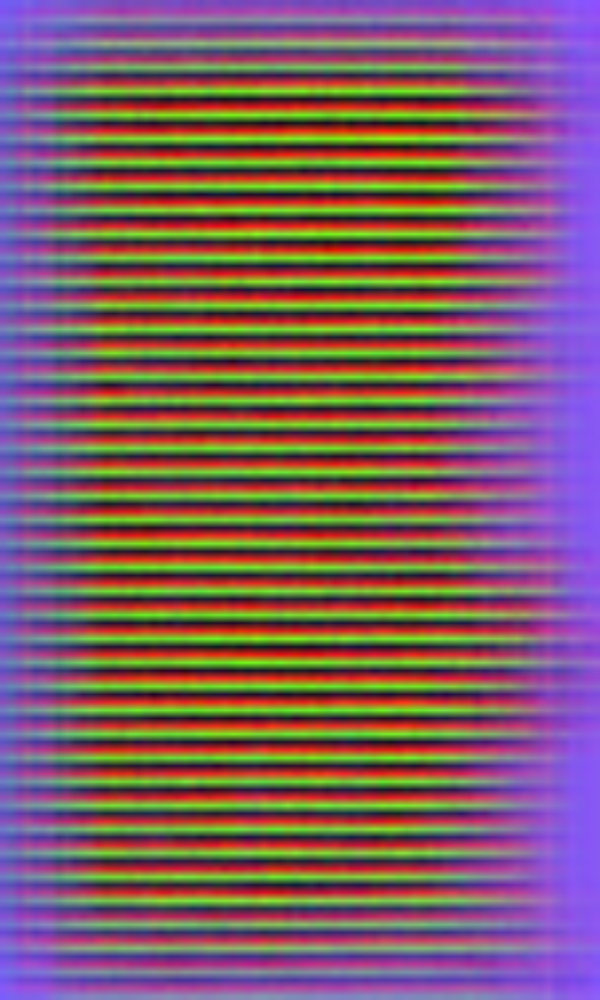} &
\includegraphics[width=0.13\textwidth, height=0.17\textwidth]{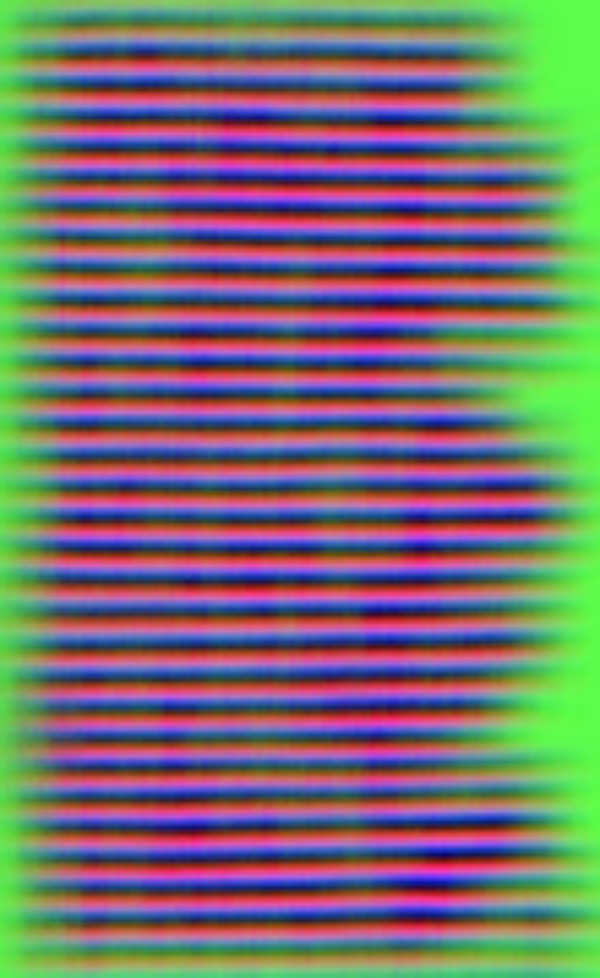} &
\includegraphics[width=0.13\textwidth, height=0.17\textwidth]{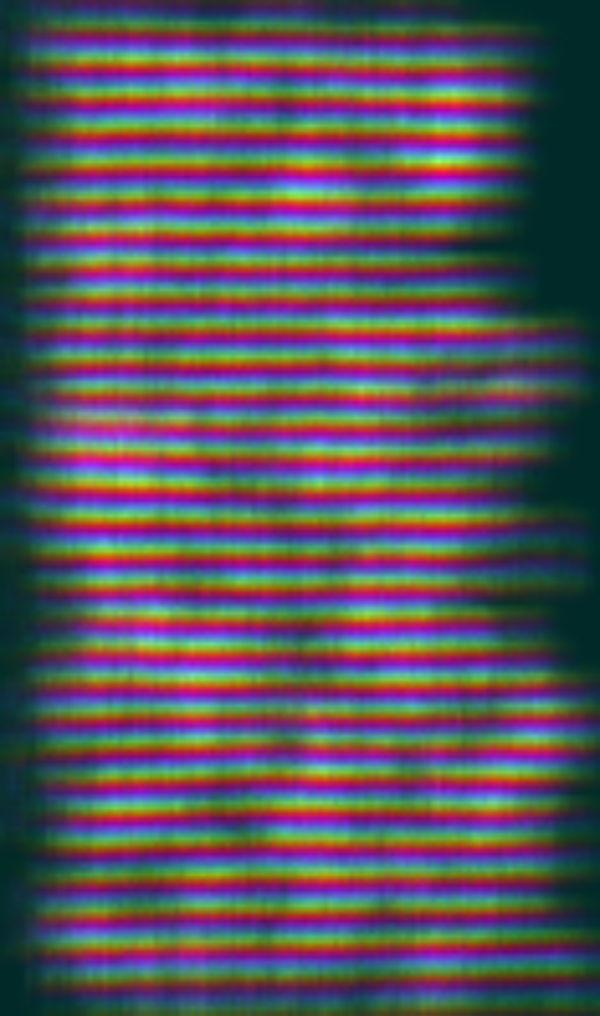} &
\includegraphics[width=0.13\textwidth, height=0.17\textwidth]{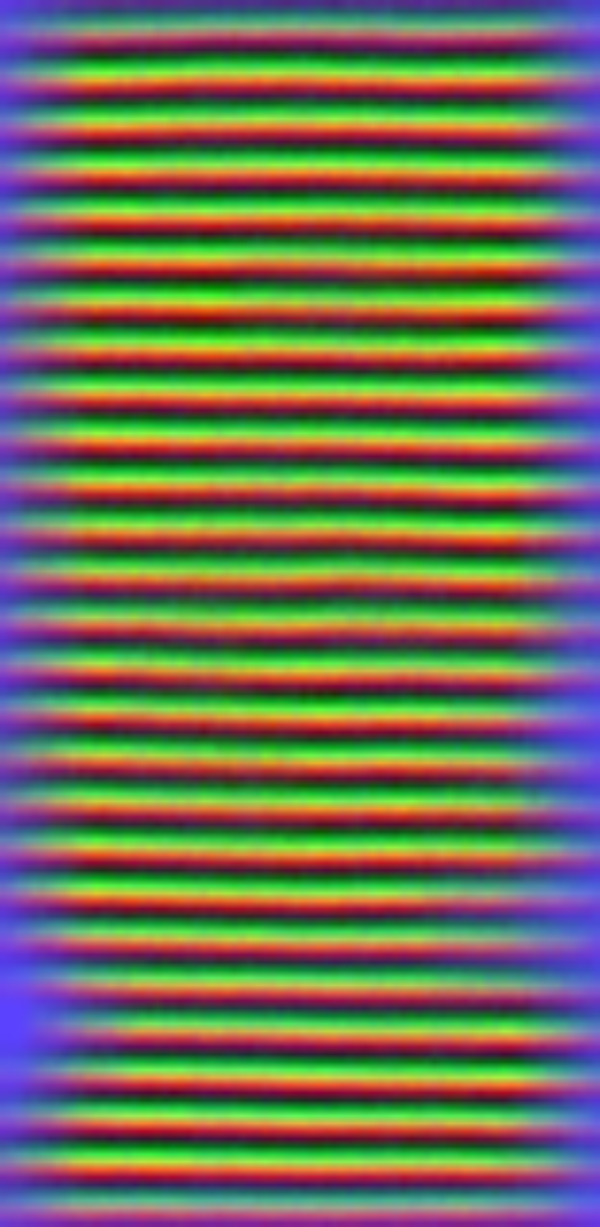}\\
\end{tabular}
\caption{The proposed method learns an embedding space in an unsupervised manner such that the distances between the embedded image patches correspond to the similarity of the coarse text line pattern they include.}
\label{fig:sample_inputs_outputs}
\end{figure*}

\section{Introduction}
\label{introduction}
Text line segmentation is a central task in document image analysis. Basically text line segmentation can be represented as text line detection and text line extraction. Text line detection is a coarse representation of text lines in terms of baselines or blob lines. Text line extraction is a fine grained representation of text lines in terms of pixel labels or bounding polygons. Once the text line detection is achieved, text line extraction is trivial using standard tools. However, text line detection is challenging due to the prevalence of irregular texture regions in handwriting.

\begin{figure*}[h]
\centering
\begin{tabular}
{m{0.19\textwidth} m{0.19\textwidth} m{0.19\textwidth} m{0.19\textwidth} m{0.19\textwidth}}
\toprule
\multicolumn{1}{m{0.19\textwidth}}{\textbf{Age}} &
\multicolumn{1}{c}{5} & \multicolumn{1}{c}{7} & \multicolumn{1}{c}{10}& \multicolumn{1}{c}{11}\\
\hline
\multicolumn{1}{m{0.19\textwidth}}{\begin{tabular}{@{}c@{}}\textbf{Perceived} \\ \textbf{text lines}\end{tabular}} &
\includegraphics[width=0.19\textwidth, height=0.19\textwidth]{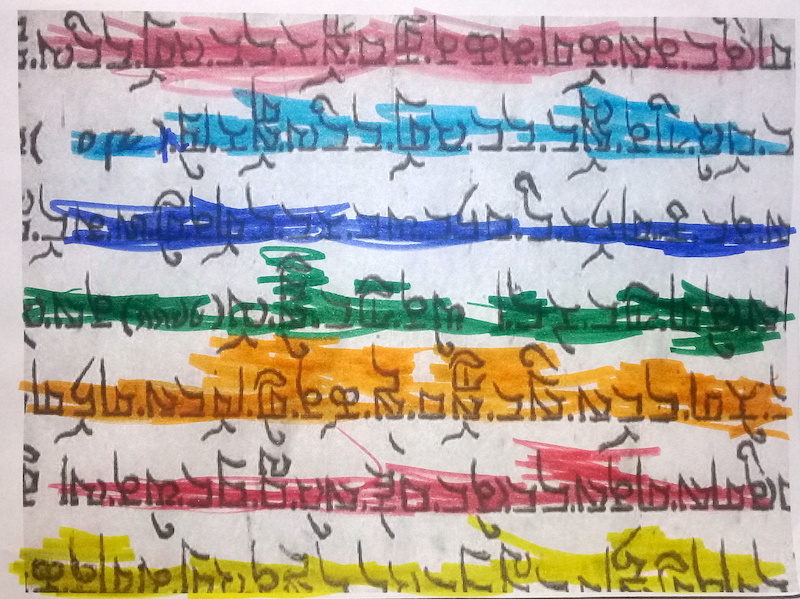} &
\includegraphics[width=0.19\textwidth, height=0.19\textwidth]{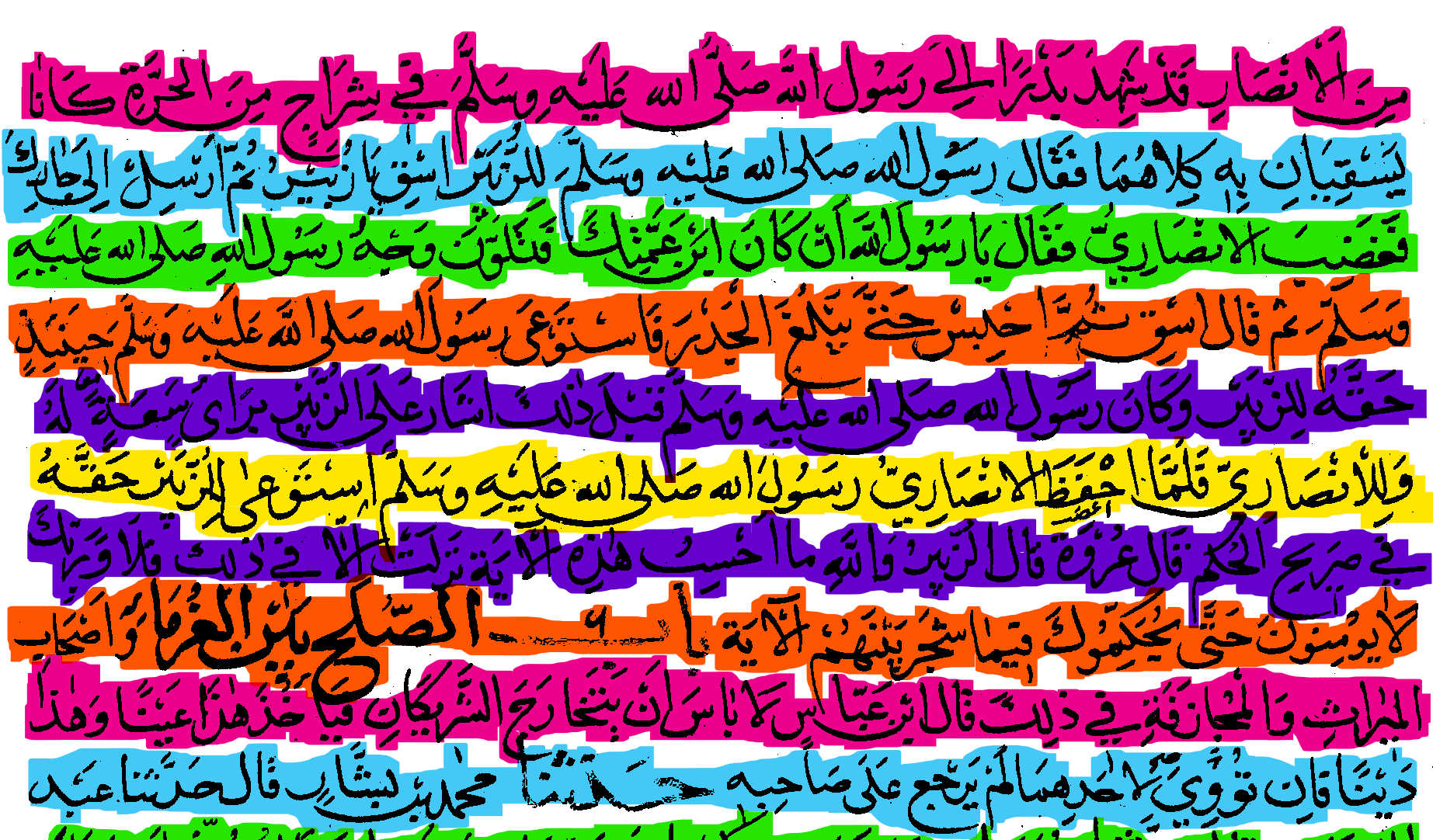} &   
\includegraphics[width=0.19\textwidth, height=0.19\textwidth]{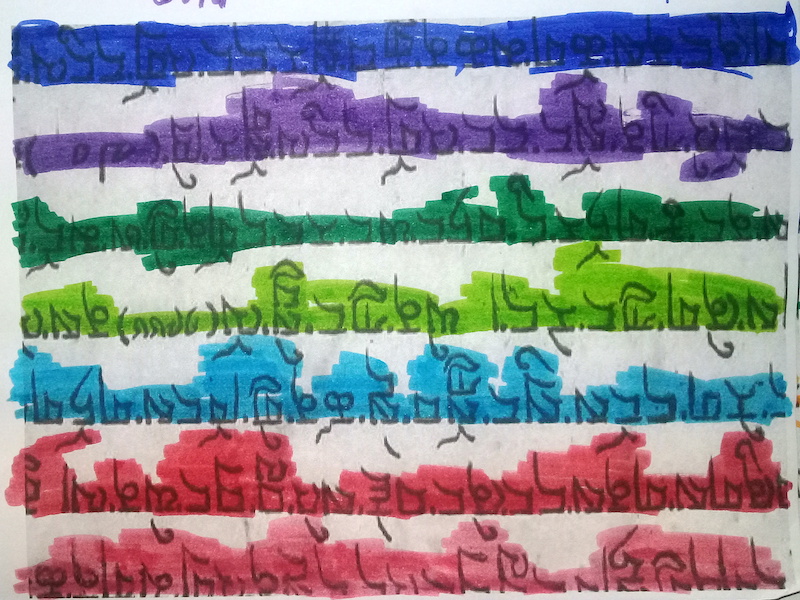} & 
\includegraphics[width=0.19\textwidth, height=0.19\textwidth]{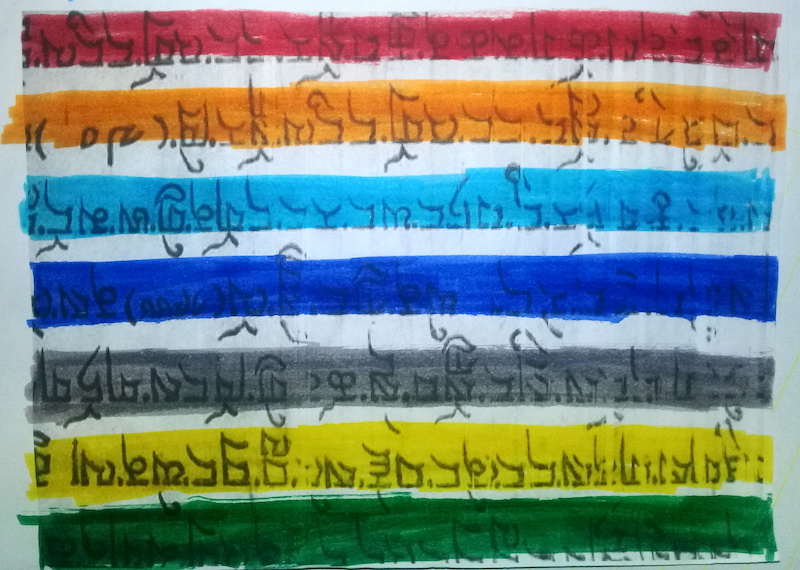} \\
\bottomrule
\end{tabular}
\caption{Human visual system can easily perceive the coarse trend of handwritten text lines. Children can segment the text lines although written in a language they are not its reader.}
\label{fig:human_visual_perception}
\end{figure*}

Given a document image patch, it contains a coarse trend of text lines. Human visual system can easily track these trend lines (\figurename~\ref{fig:human_visual_perception}), but a computer algorithm cannot track them due to the textured structure of each text line at fine details. Inspired by this fact, we hypothesize that a convolutional network can be trained in an unsupervised manner to map document image patches to some vector space such that the patches with the same coarse text line pattern are proximate and the patches with different coarse text line pattern are distant. We can assume that two neighbouring patches contain the same coarse text line pattern and contain different coarse text line pattern if one of them is rotated 90 degrees. Doing well on this task requires the model to learn to recognize the text lines and their salient parts. Hence the embedded features of document patches can also be used to discriminate the differences in the horizontal text line patterns that they contain. 
Clustering the patches of a document page by projecting their vectors onto three principle directions yields a pseudo-rgb image where coarse text line patterns correspond to similar colours (\figurename~\ref{fig:sample_inputs_outputs}). The pseudo-rgb image can then be thresholded into blob lines that strikethrough the text lines and guide an energy minimization function for extracting the text lines. 

The proposed method has been evaluated on two publicly available handwritten documents dataset. The results demonstrate that this unsupervised learning method provides interesting text line segmentation results on handwritten document images. 

\section{Related work}
The recent trend in solving the handwritten text line segmentation problem is to employ deep networks that learn the representation directly from the pixels of the image rather than using engineered features \cite{diem2017cbad}. These methods use a large dataset of labelled text lines to acquire the texture variances due to different font types, font sizes, and orientations. 

Early attempts formulate text line segmentation as a supervised binary dense prediction problem. Given a document image, a Fully Convolutional Network (FCN) \cite{long2015fully} is trained to densely predict whether a pixel is a text line pixel or not. However, the question that arises here is: Which pixels belong to a text line? Foreground pixels definitely can not discriminate a text line from the others because FCN output is a semantic segmentation where multiple instances of the same object are not separated. Very recently, text line segmentation has been formulated as an instance segmentation problem using Mask-RCNN \cite{he2017mask}, and its results are available in \cite{kurar2020text}. However, when using FCN, each text line is represented as a single connected component. This component can be either a blob line \cite{vo2017text,renton2018fully,kurar2018text,mechi2019text,kurar2020text} strikes through the main body area of the characters that belong to a text line or a baseline \cite{gruning2019two} passes through the bottom part of the main body of the characters that belong to a text line. FCNs are very successful at detecting handwritten text lines \cite{diem2017cbad}. However, scarcity of labelled data causes rarely occurring curved text lines to be poorly detected. This problem has been handled via augmentation \cite{gruning2019two} or learning-free detection \cite{kurar2019vml}. 

Both the text line representations, blob line and baseline, are coarse grained representations and do not fully label all the pixels of a text line but only detect the spatial location of a text line. There are metrics that can evaluate the detected baselines~\cite{diem2017cbad,renton2018fully,mechi2019text} or blob lines \cite{kurar2018text}. Alternatively, detected spatial location of text lines are utilized to further extract the pixels of text lines. Some of these extraction methods assume horizontal text lines~\cite{vo2017text} whereas some can extract text lines at any orientation, with any font type and font size~\cite{kurar2020text}. Text line extraction is evaluated by classical image segmentation metrics~\cite{diem2017cbad}.

Deep networks have apparently increased handwritten text line segmentation performance by their ability to learn comprehensive visual features. However, they need to leverage large labelled datasets, which in turn brings costly human annotation effort. Learning-free algorithms would be a natural solution but still they do not achieve state of the art ~\cite{kurar2020learning} except used in hybrid with deep networks ~\cite{alberti2019labeling}. Another solution would be unsupervised learning methods. However, the main concern is to find an objective function that will use a representation to capture text lines, although they are not labelled. Kurar~\etal~\cite{kurar2020unsupervised} formulated this concern as the answer to the question of whether a given document patch contains a text line or space line. The answer is based on a human adjusted score. In this paper, we propose an unsupervised text line segmentation method that trains a deep network to answer whether two document image patches contain the same coarse text line pattern or different coarse text line pattern. The network is urged to learn the salient features of text lines in order to answer this question.

\section{Method}
\label{method}

\begin{figure}[t]
\centering
\includegraphics[width=0.95\textwidth]{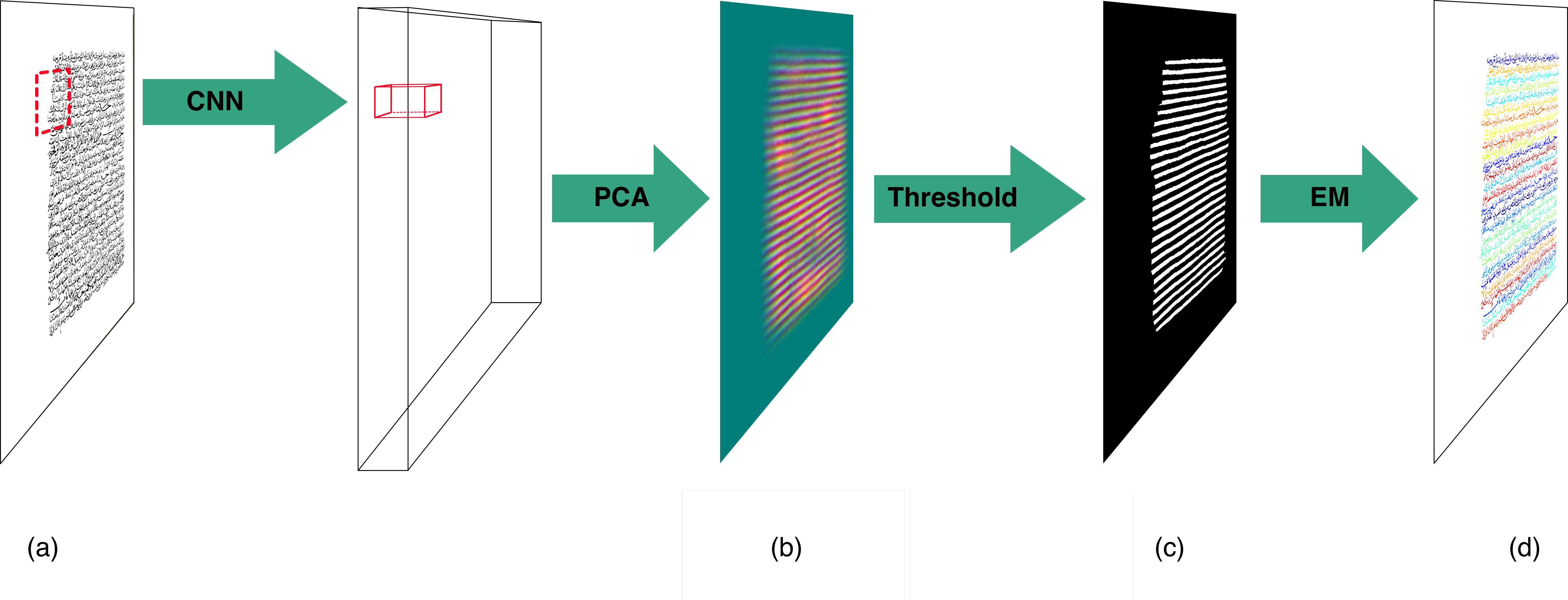}
\caption{Given a handwritten document image (a), first stage extracts feature vectors of image patches such that the patches with similar text line trends are close in the space. Second stage clusters the patches of a document image according to the first three principal components of their feature vectors. This stage outputs the a pseudo-rgb image (b) which is then thresholded onto blob lines (c) that strike through text lines. Energy minimization with the assistance of detected blob lines extracts the pixel labels of text lines (d).}
\label{fig:stages}
\end{figure}

Unsupervised learning of text line segmentation is a three stage method (\figurename~\ref{fig:stages}). The first stage relies on a deep convolutional network that can predict a relative similarity for a pair of patches and embed the patches into feature vectors. The similarity of two patches in document images correlates with their text line orientation assuming that the neighbouring patches contain the same orientation. The second stage generates a pseudo-rgb image using the three principals of the feature vectors obtained from the first stage. The pseudo-rgb image is further thresholded to detect the blob lines that strike through the text lines. Final stage performs pixel labelling for text lines using an energy minimization function that is assisted by the detected blob lines.

\subsection{Deep convolutional network}
Convolutional networks are well known to learn complex image representations from raw pixels. We aim the convolutional network to learn the coarse trend of text lines. We train it to predict the similarity for a pair of patches in terms of text line orientation. In a given document image neighbouring patches would contain the same coarse trend of text lines. Therefore, the network is expected to learn a feature embedding such that the patches that contain the same text line pattern would be close in the space. 

To achieve this we use a pair of convolutional networks with shared weights such that the same embedding function is computed for both patches. Each convolutional branch processes only one of the patches hence the network performs most of the semantic reasoning for each patch separately. Consequently, the feature representations are concatenated and fed to fully connected layers in order to predict whether the two image patches are similar or different. 

\begin{figure}[t]
\centering
\includegraphics[width=7cm]{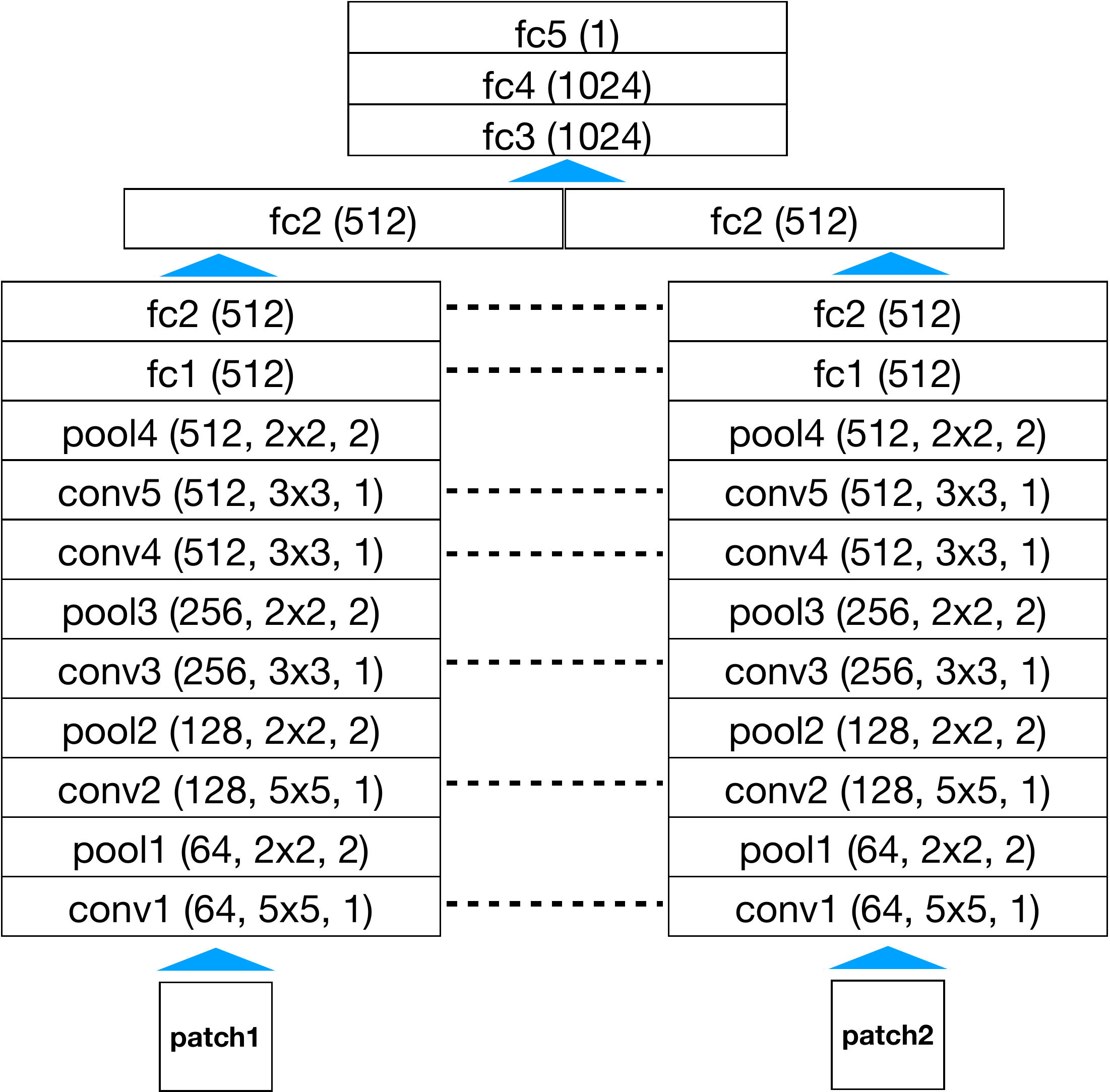}
\caption{Convolutional network architecture for pair similarity. Dotted lines stand for identical weights, conv stands for convolutional layer, fc stands for fully connected layer and pool is a max pooling layer.}
\label{fig:siamese}
\end{figure}

The architecture of the branches is based on AlexNet \cite{krizhevsky2012imagenet} and through experiments we tune the hyperparameters to fit our task. Each of the branches has five convolutional layers as presented in \figurename~\ref{fig:siamese}. Dotted lines indicate identical weights, and the numbers in parentheses are the number of filters, filter size and stride. All convolutional and fully connected layers are followed by ReLU activation functions, except fc5, which feeds into a sigmoid binary classifier.

\subsubsection{Pair generation}
\begin{figure}[t]
\centering
\includegraphics[width=0.95\textwidth]{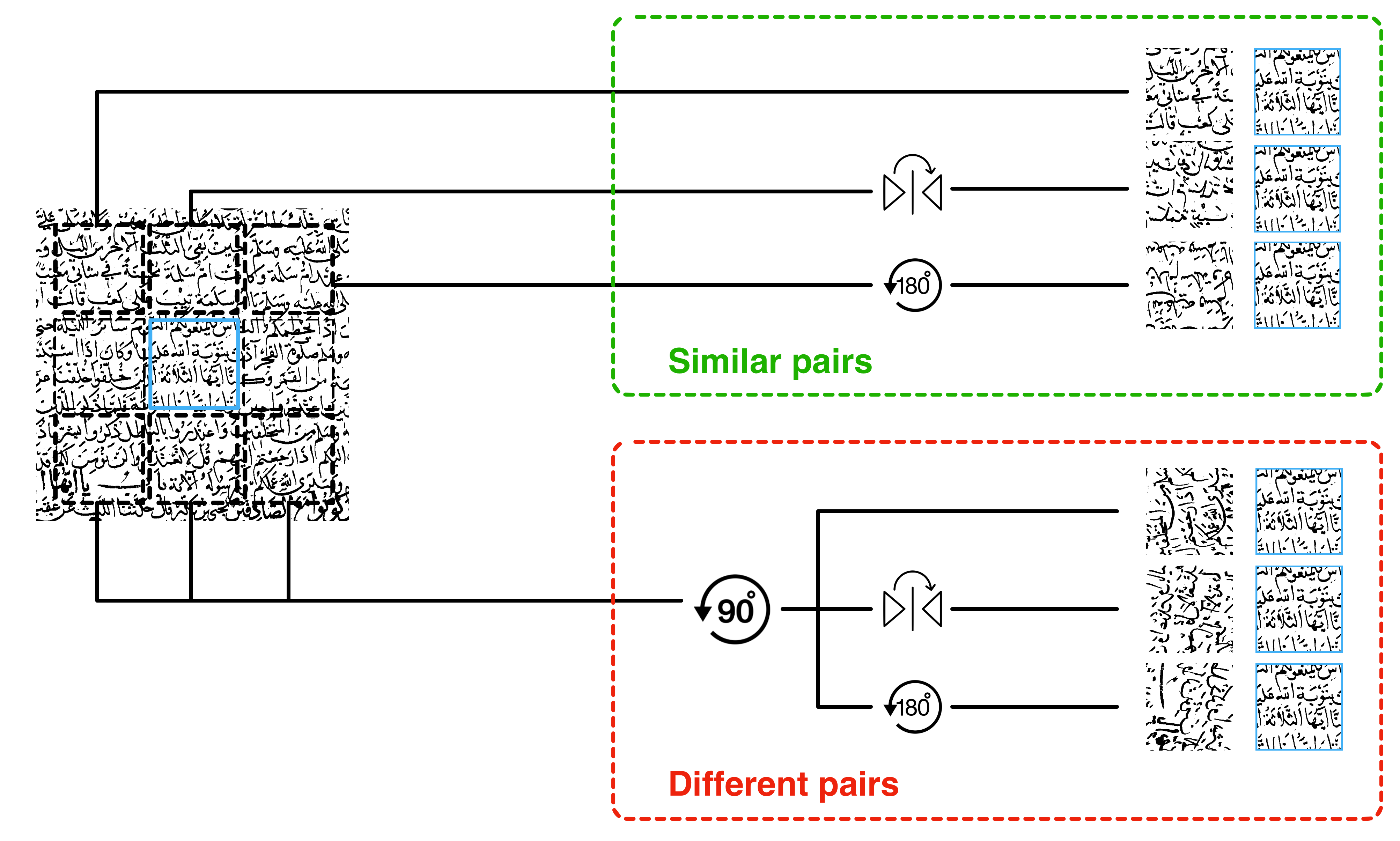}
\caption{The pairs are generated with the assumption that neighbouring patches contain similar text line trends. Different pairs are generated by rotating one of the patches 90 degrees. Both, the similar and different, pairs are augmented by randomly rotating one of the patches $0$ degrees or $180$ degrees or flipping.}
\label{fig:pair_generation}
\end{figure}

Given a document image, we sample the first patch uniformly from regions containing foreground pixels. Given the position of the first patch we sample the second patch randomly from the eight possible neighbouring locations. We include a gap and jitter between patches in order to prevent cues like boundary patterns or lines continuing between patches. Neighbouring patches in a document image can be assumed to contain the same text line orientation and are labeled as similar pairs. Different pairs are generated by rotating the second patch $90$ degrees. Additionally for both, the similar pairs and the different pairs, the second patches are randomly rotated $0$ degrees or rotated $180$ degrees or flipped. Pair generation is demonstrated in \figurename~\ref{fig:pair_generation}. In case of fluctuating or skewed text lines, the similarity does not correlate with the proximity. However in a document image with almost all horizontal text lines these dissimilar and close patches are rare.

\subsubsection{Training}
For each dataset we train the model from scratch using $n_p$ pairs:
\begin{equation}
    n_p = \frac{h_a \times w_a}{p\times p} \times n_d
\label{eq:number_of_patches}
\end{equation}
where $h_a$ and $w_a$ are the average document image height and width in the dataset, $p$ is the patch size, and $n_d$ is the number of document images in the set. The learning rate is $0.00001$, the batch size is $8$ and the optimizing algorithm is Adam. We continue training until there is no improvement on the validation accuracy with a patience of $7$ epochs and save the model with the best validation accuracy for the next stage.

\subsection{Pseudo-rgb image}
The convolutional network performs most of the semantic reasoning for each patch separately because only three layers receive input from both patches. Hence we can use a single branch to extract the significant features of patches. This embeds every patch into a feature vector of $512$ dimensions. To visualize the features of a complete document image, a sliding window of the size $p\times p$ is used, but only the inner window of the size $w\times w$ is considered to increase the representation resolution. We also pad the document image with background pixels at its right and bottom sides if its size is not an integer multiple of the sliding window size. An additional padding is added at four sides of the document image for considering only the central part of the sliding window. Resultantly, a document image with the size $h_d\times w_d$ is mapped to a representation matrix of the size $\frac{h_d}{w}\times \frac{w_d}{w}\times 512$. We project $512D$ vectors into their three principle components and use these components to construct pseudo-rgb image in which similar patches are assigned the similar colors (\figurename~\ref{fig:stages}(b)). Binary blob lines image is an outcome of thresholded pseudo-rgb image (\figurename~\ref{fig:stages}(c)).

\subsection{Energy minimization}
We adopt the energy minimization framework \cite{boykov2001fast} that uses graph cuts to approximate the minimal of an arbitrary function. We adapt the energy function to be used with connected components for extracting the text lines. Minimum of the adapted function correspond to a good extraction which urges to assign components to the label of the closest blob line while straining to assign closer components to the same label (\figurename~\ref{fig:stages}(d)). A touching component $c$ among different blob lines is split by assigning each pixel in $c$ to the label of the closest blob line.

Let $\mathcal{L}$ be the set of binary blob lines, and $\mathcal{C}$ be the set of components in the binary document image. Energy minimization finds a labeling $f$ that assigns each component $c\in \mathcal{C}$ to a label $l_c\in \mathcal{L}$, where energy function $\textbf{E}(f)$ has the minimum.

\begin{equation}
    \textbf{E}(f) = \sum_{c\in {\mathcal C}}D(c, \ell_c)+\sum_{\{c,c'\}\in \mathcal N}d(c, c')\cdot \delta (\ell_c \neq \ell_{c'})
\label{eq:em}
\end{equation}

The term $D$ is the data cost, $d$ is the smoothness cost, and  $\delta$ is an indicator function. Data cost is the cost of assigning component $c$ to label $l_c$. 
$D(c, \ell_c)$ is defined to be the Euclidean distance between the centroid of the component $c$ and the nearest neighbour pixel in blob line $l_c$ for the centroid of the component $c$. 
Smoothness cost is the cost of assigning neighbouring elements to different labels. Let $\mathcal{N}$ be the set of nearest component pairs. Then $\forall \{c,c'\}\in \mathcal {N}$

\begin{equation}
    d(c,c') = \exp({-\beta\cdot d_c(c,c')})
\label{eq:sc}
\end{equation}
where $d_c(c,c')$ is the Euclidean distance between the centroids of the components $c$ and $c'$,  and $\beta$ is defined as
\begin{equation}
     \beta=(2\left<d_c(c,c')\right>)^{-1}
\end{equation}
$\left<\cdot\right>$ denotes expectation over all pairs of neighbouring components \cite{boykov2001interactive} in a document page image. $\delta (\ell_c \neq \ell_{c'})$ is equal to $1$ if the condition inside the parentheses holds and $0$ otherwise.

\section{Experiments}
\label{experiments}

In this section we first introduce the datasets used in the experiments. We define the parameters of the baseline experiment, and investigate the influence of patch size and central window size on the results. Then we visualize patch saliency for understanding the unsupervised learning of text line segmentation. Finally we discuss the limitations of the method.

\subsection{Data}
The experiments cover five datasets that are different in terms of the challenges they pose. The VML-AHTE dataset \cite{kurar2020text} consists of Arabic handwritten documents with crowded diacritics and cramped text lines. The Pinkas dataset \cite{barakat2019pinkas} contains slightly edge rounded and noisy images of Hebrew handwritten documents. Their ground truth is provided in PAGE xml format \cite{pletschacher2010page,clausner2011aletheia}. The Printed dataset is our private and synthetic dataset that is created using various font types and sizes. The ICFHR2010 \cite{gatos2010icfhr} is a dataset of modern handwriting that is heterogeneous  by document resolutions, text line heights and skews. The ICDAR2017 dataset \cite{simistira2017icdar2017} includes three books, CB55, CSG18, and CSG863. In this dataset we run our algorithm on presegmented main text regions by the given ground truth. The VML-MOC dataset \cite{barakat2019vml} is characterized by multiply oriented and curved handwritten text lines.

\begin{figure}[t]
\centering
\includegraphics[width=10cm]{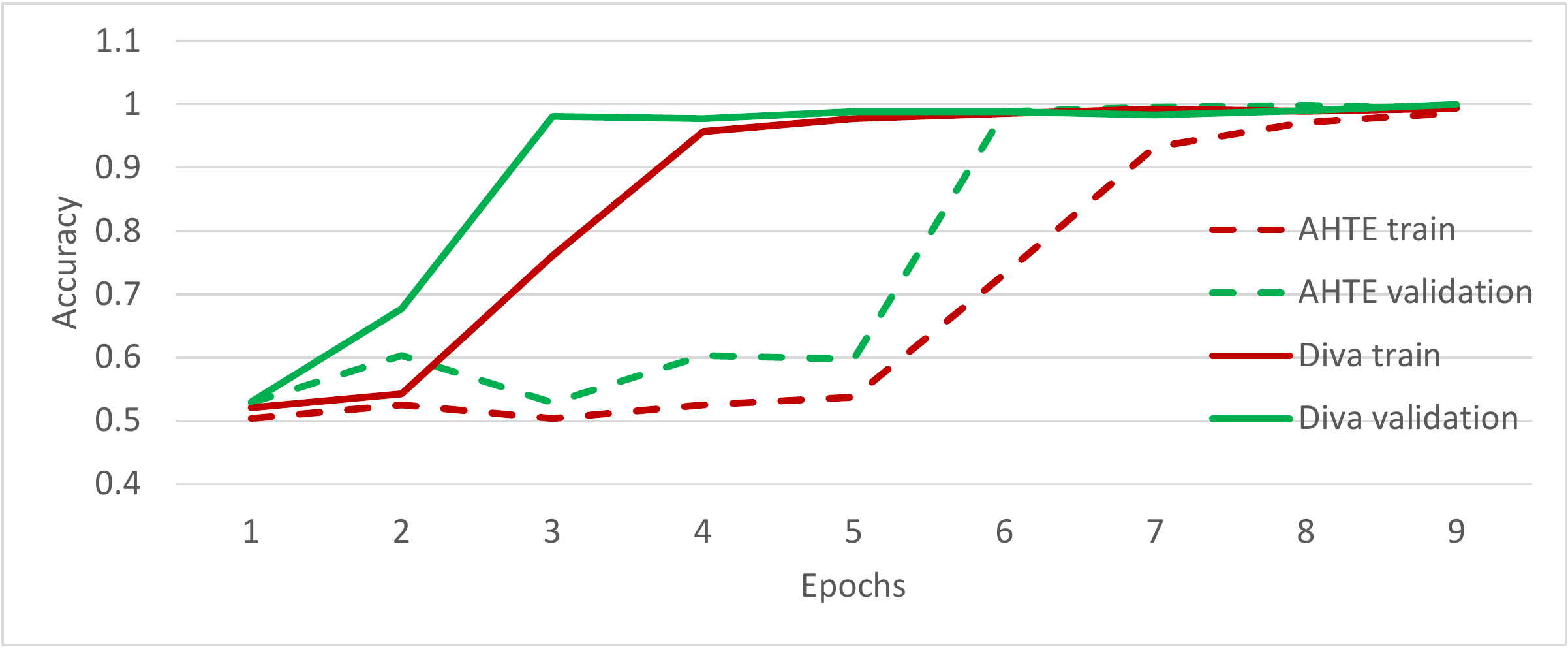}
\caption{Train and validation logs on the VML-AHTE and ICDAR2017 datasets.}
\label{fig:trainlog}
\end{figure}

\subsection{Baseline experiment}
\begin{figure*}[t]
\centering
\begin{tabular}
{m{0.22\textwidth} m{0.22\textwidth} m{0.22\textwidth} m{0.22\textwidth}}

\multicolumn{1}{c}{Pinkas} & \multicolumn{1}{c}{ICFHR} & \multicolumn{1}{c}{AHTE} & \multicolumn{1}{c}{MOC} \\

\includegraphics[width=0.22\textwidth, height=0.33\textwidth]{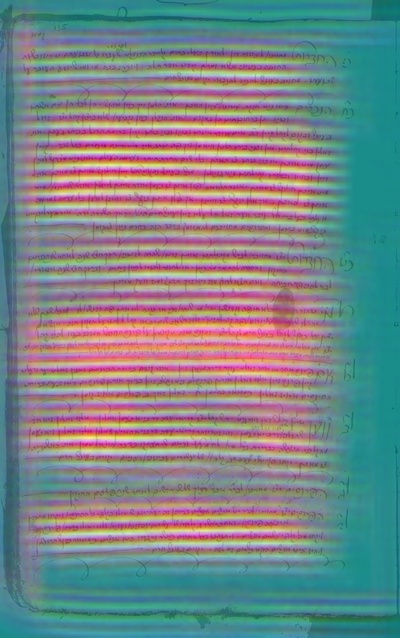} &
\includegraphics[width=0.22\textwidth, height=0.33\textwidth]{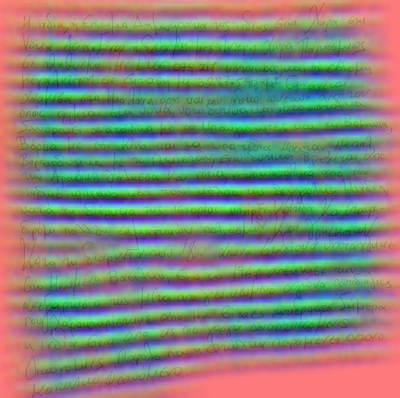} & 
\includegraphics[width=0.22\textwidth, height=0.33\textwidth]{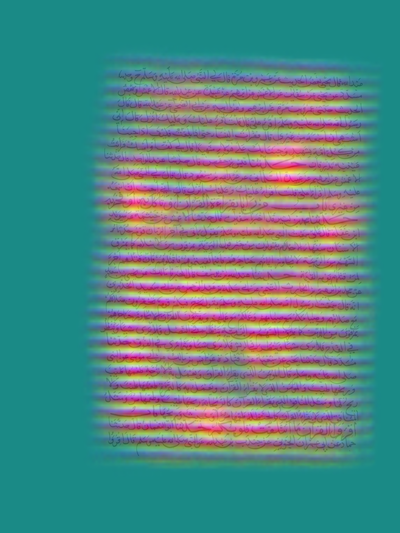} &
\includegraphics[width=0.22\textwidth, height=0.33\textwidth]{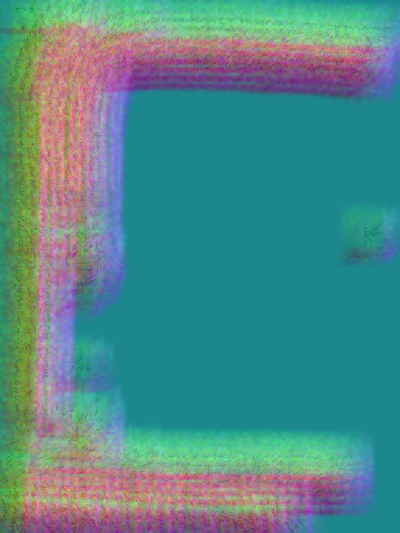} \\

\multicolumn{1}{c}{Printed} & \multicolumn{1}{c}{CSG-863} & \multicolumn{1}{c}{CB-55} & \multicolumn{1}{c}{CSG-18}\\

\includegraphics[width=0.22\textwidth, height=0.33\textwidth]{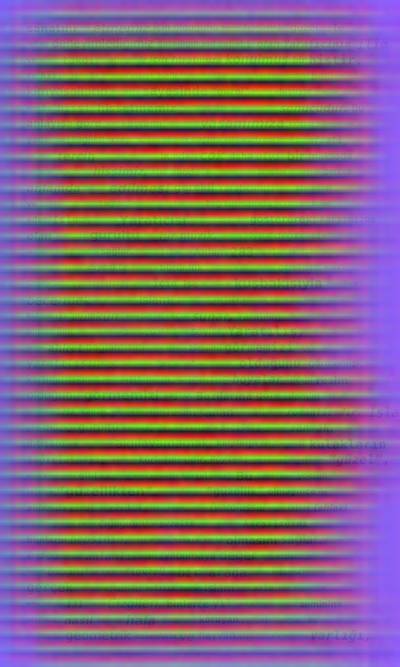} &
\includegraphics[width=0.22\textwidth, height=0.33\textwidth]{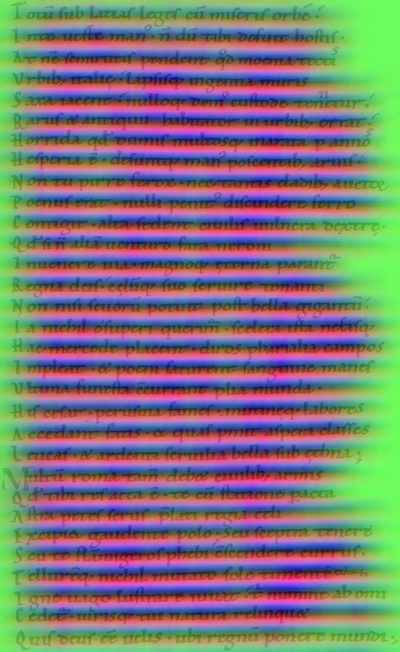} &
\includegraphics[width=0.22\textwidth, height=0.33\textwidth]{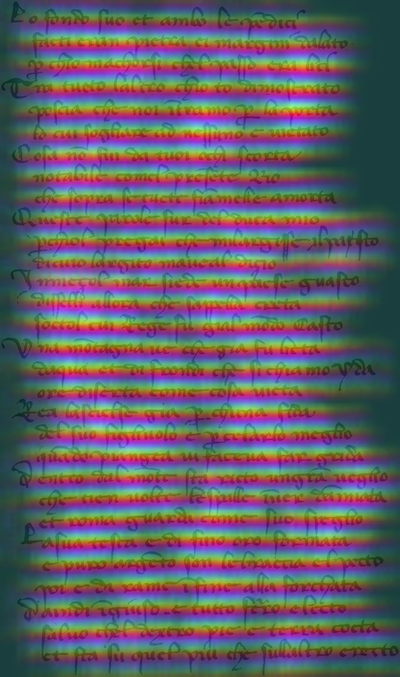} &
\includegraphics[width=0.22\textwidth, height=0.33\textwidth]{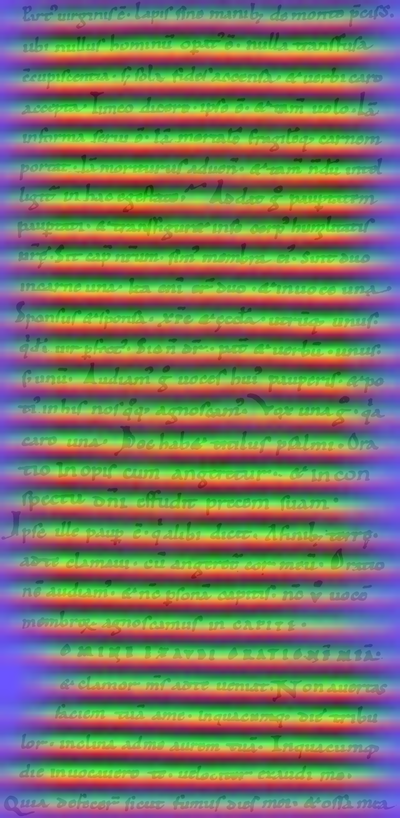}\\
\end{tabular}
\caption{The results of baseline experiment are shown overlapped with the input images. The result on the VML-MOC dataset is a mess because the method assumes almost horizontal text lines when labeling the similar and different pairs.}
\label{fig:baseline_results}
\end{figure*}

We choose to experiment on five datasets with different challenges in order to verify that the method generalizes. Therefore, we define a baseline experiment that set the parameter values. There is no best set of parameters that fit all challenges and one can always boost the performance on a particular dataset by ad-hoc adjusting. However we wish to propose a baseline experiment that can fit all challenges as much as possible. The baseline experiment sets the input patch size $p=350$, and the sliding central window size $w=20$. The results are shown in \figurename~\ref{fig:baseline_results}. The convolutional network easily learns the embedding function. The validation accuracy almost always reaches over $99\%$ (\figurename~\ref{fig:trainlog}). We have preliminary experiment which suggest that increasing the number of layers until VGG-16 and then until VGG-19 leads to successful blob detection as well as AlexNet do. However, a deeper network such as ResNet does not detect blobs, probably because the reception field of its last convolutional layer is larger.

\subsection{Effect of patch size ($p$)}
\begin{figure*}[ht]
\centering
\begin{tabular}
{m{0.018\textwidth} m{0.13\textwidth} m{0.13\textwidth} m{0.13\textwidth} m{0.13\textwidth} m{0.13\textwidth} m{0.13\textwidth} m{0.13\textwidth}}

\multicolumn{1}{c}{} & \multicolumn{1}{c}{Pinkas} & \multicolumn{1}{c}{ICFHR} & \multicolumn{1}{c}{AHTE} & \multicolumn{1}{c}{Printed} & \multicolumn{1}{c}{CSG-863} & \multicolumn{1}{c}{CB-55} & \multicolumn{1}{c}{CSG-18}\\

\rotatebox[origin=c]{90}{400} &
\includegraphics[width=0.13\textwidth, height=0.19\textwidth]{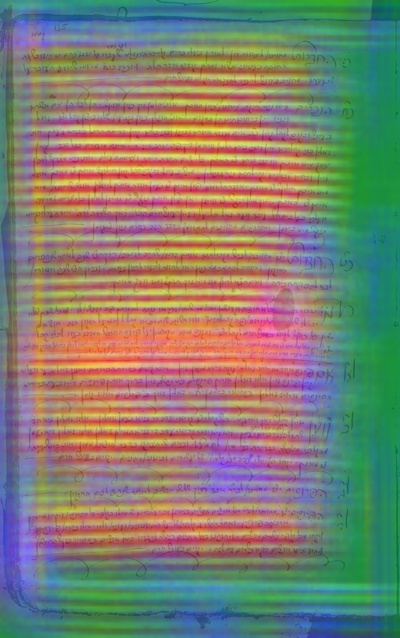} &
\includegraphics[width=0.13\textwidth, height=0.19\textwidth]{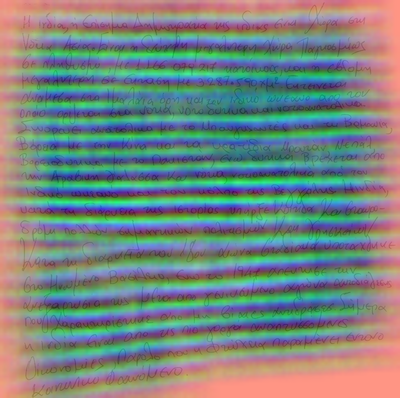} & 
\includegraphics[width=0.13\textwidth, height=0.19\textwidth]{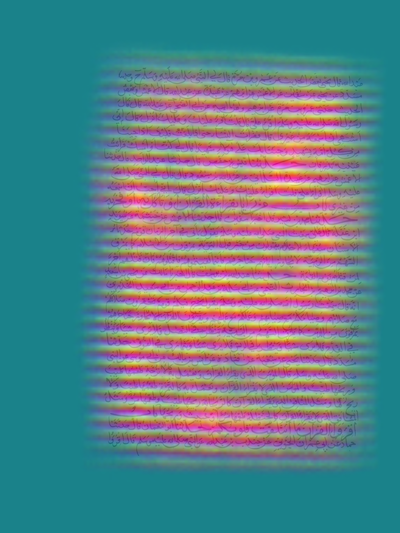} &
\includegraphics[width=0.13\textwidth, height=0.19\textwidth]{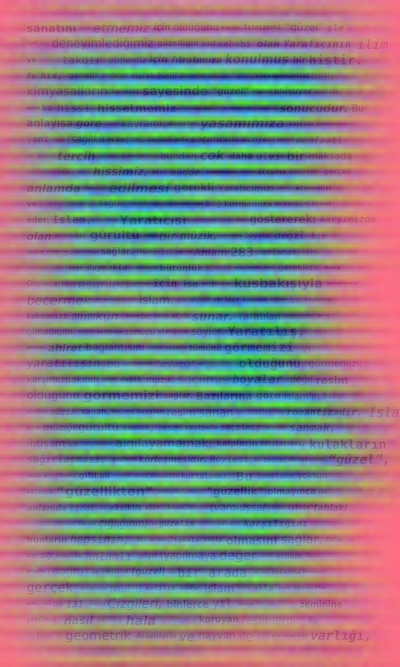} &
\includegraphics[width=0.13\textwidth, height=0.19\textwidth]{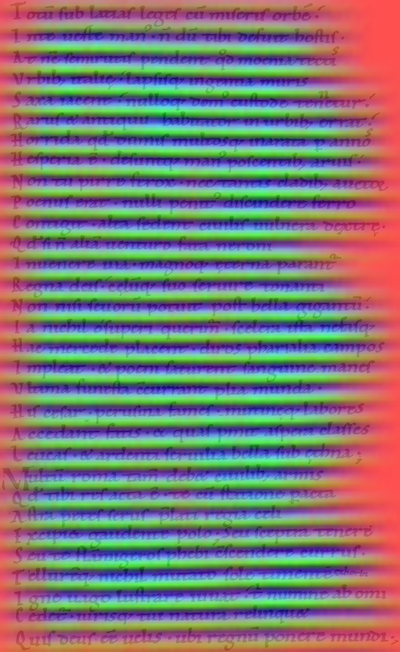} &
\includegraphics[width=0.13\textwidth, height=0.19\textwidth]{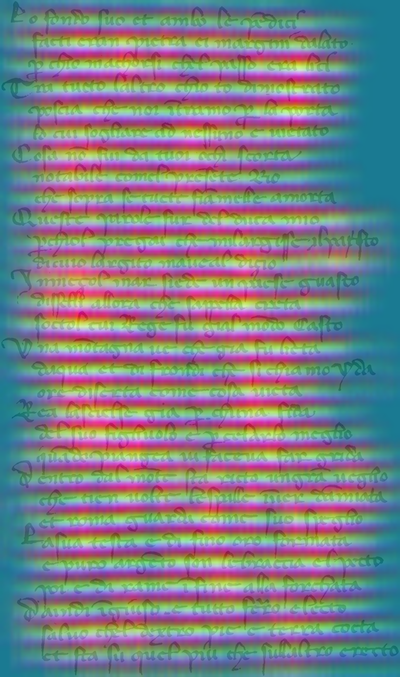} &
\includegraphics[width=0.13\textwidth, height=0.19\textwidth]{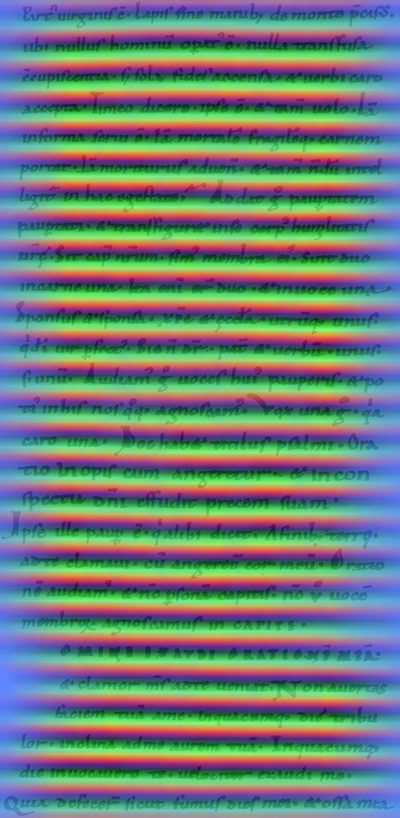}\\

\rotatebox[origin=c]{90}{300} &
\includegraphics[width=0.13\textwidth, height=0.19\textwidth]{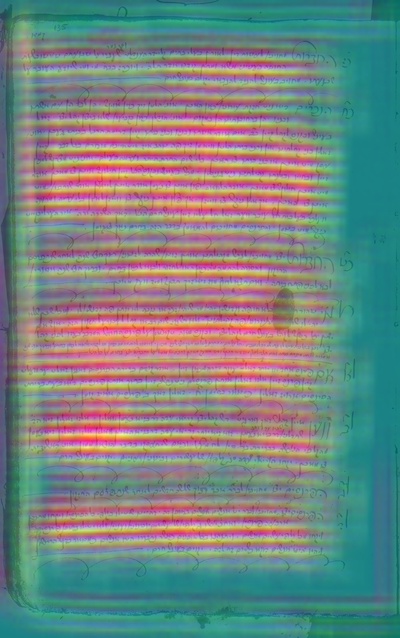} &
\includegraphics[width=0.13\textwidth, height=0.19\textwidth]{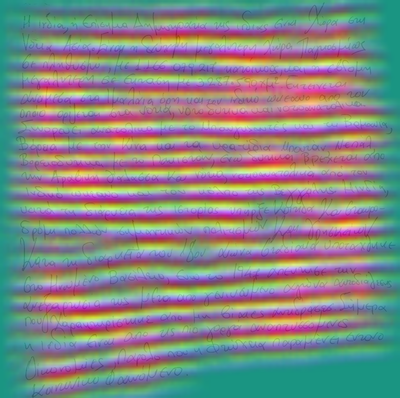} & 
\includegraphics[width=0.13\textwidth, height=0.19\textwidth]{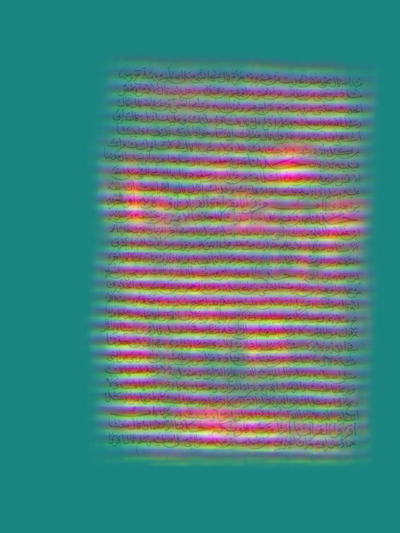} &
\includegraphics[width=0.13\textwidth, height=0.19\textwidth]{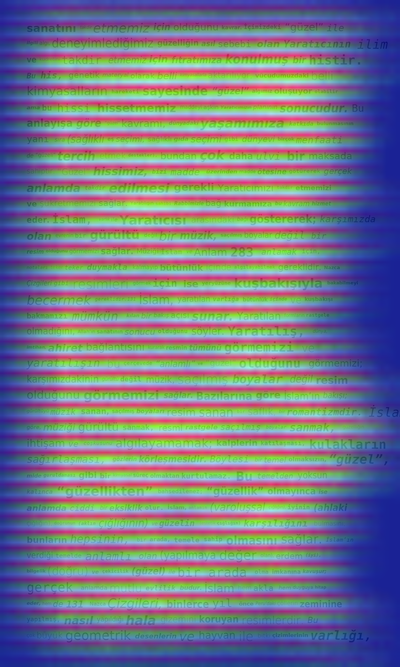} &
\includegraphics[width=0.13\textwidth, height=0.19\textwidth]{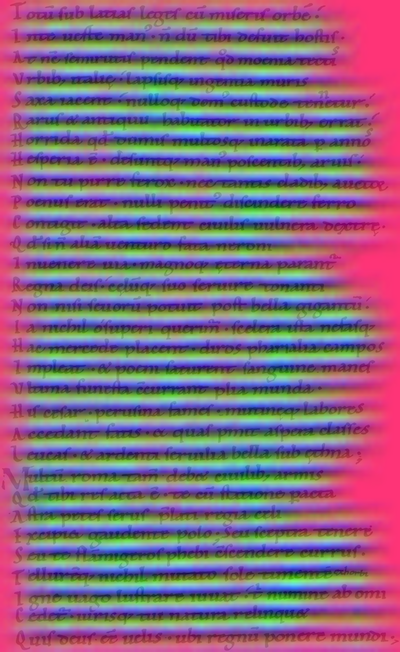} &
\includegraphics[width=0.13\textwidth, height=0.19\textwidth]{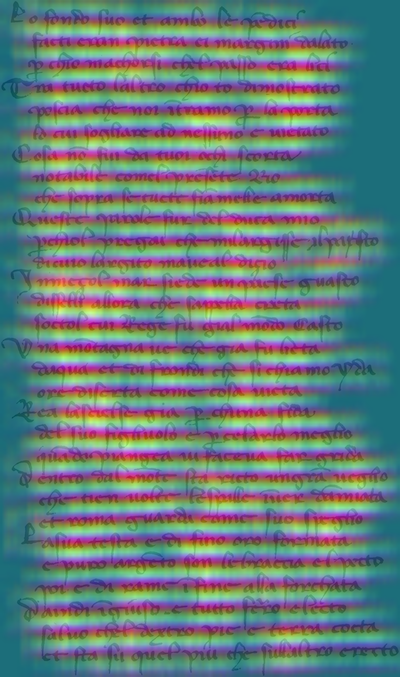} &
\includegraphics[width=0.13\textwidth, height=0.19\textwidth]{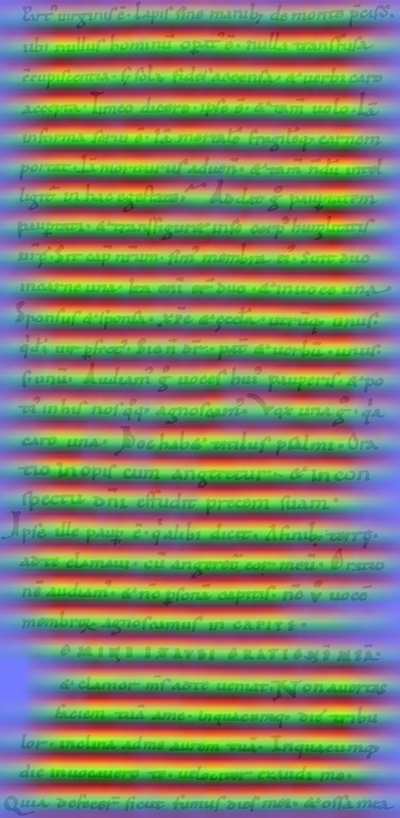}\\

\rotatebox[origin=c]{90}{200} &
\includegraphics[width=0.13\textwidth, height=0.19\textwidth]{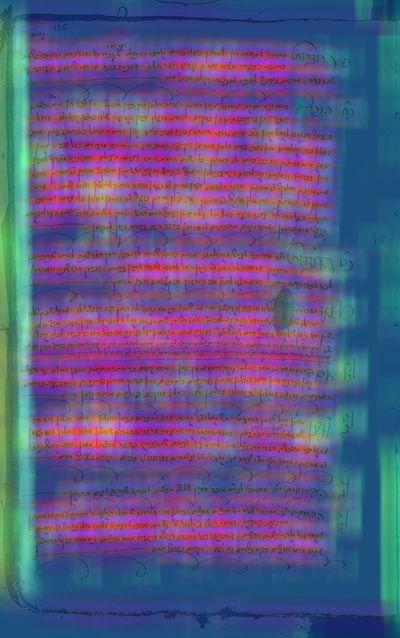} &
\includegraphics[width=0.13\textwidth, height=0.19\textwidth]{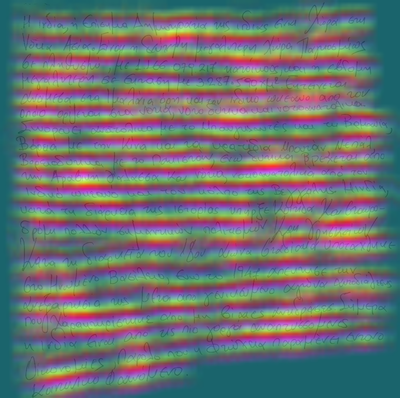} & 
\includegraphics[width=0.13\textwidth, height=0.19\textwidth]{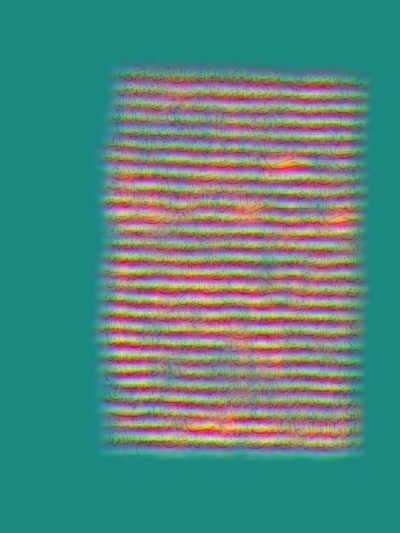} &
\includegraphics[width=0.13\textwidth, height=0.19\textwidth]{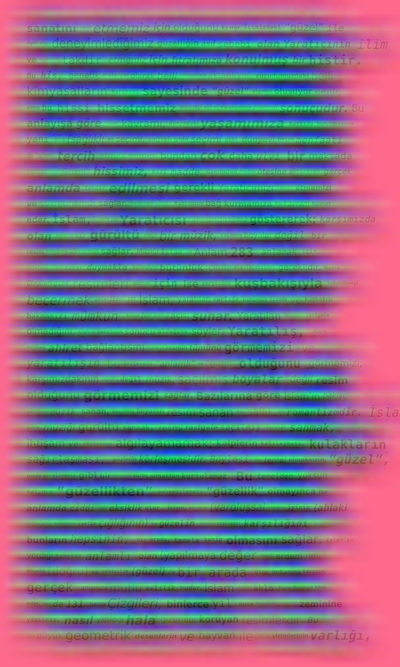} &
\includegraphics[width=0.13\textwidth, height=0.19\textwidth]{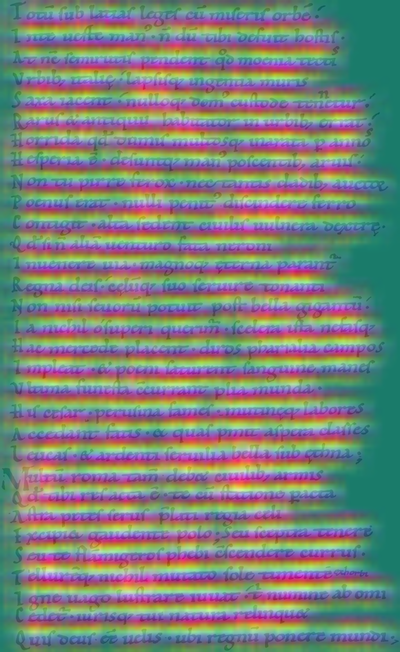} &
\includegraphics[width=0.13\textwidth, height=0.19\textwidth]{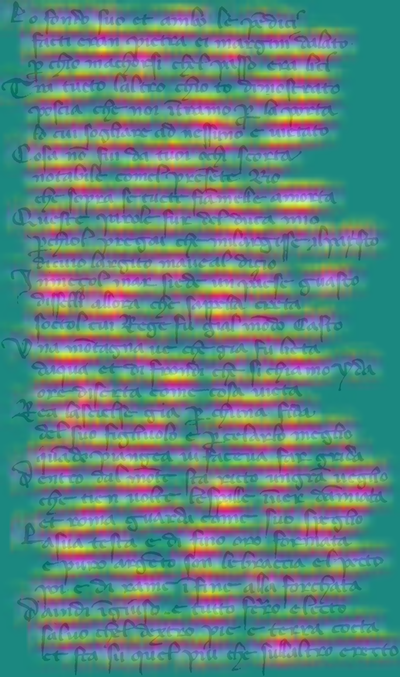} &
\includegraphics[width=0.13\textwidth, height=0.19\textwidth]{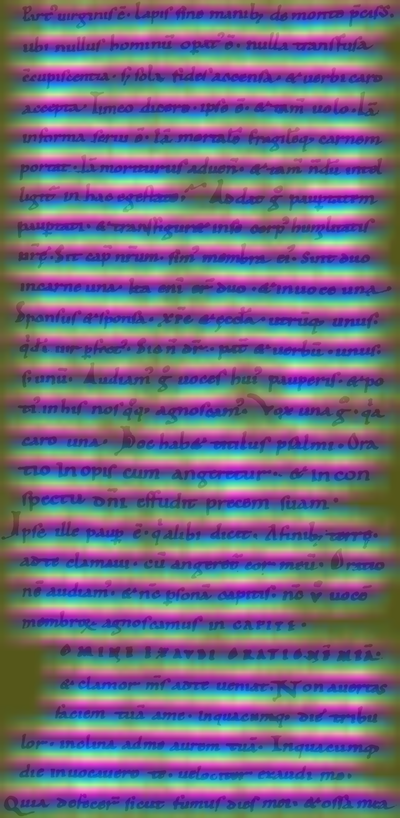}\\

\rotatebox[origin=c]{90}{100} &
\includegraphics[width=0.13\textwidth, height=0.19\textwidth]{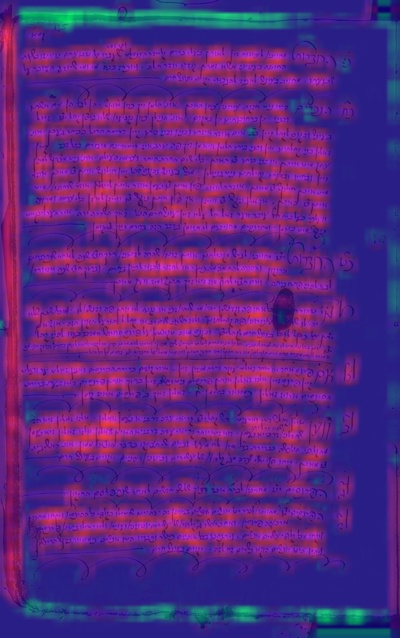} &
\includegraphics[width=0.13\textwidth, height=0.19\textwidth]{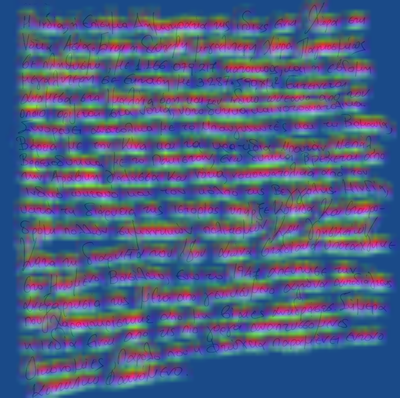} & 
\includegraphics[width=0.13\textwidth, height=0.19\textwidth]{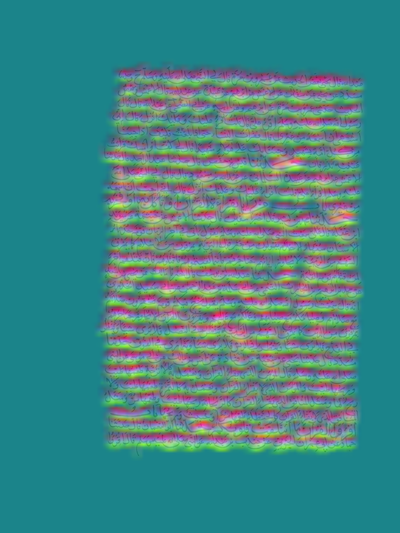} &
\includegraphics[width=0.13\textwidth, height=0.19\textwidth]{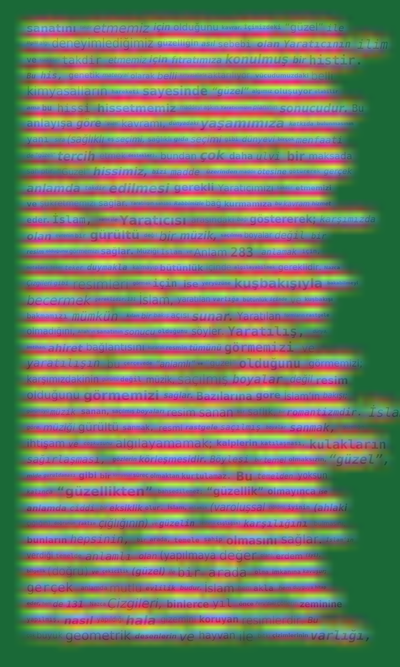} &
\includegraphics[width=0.13\textwidth, height=0.19\textwidth]{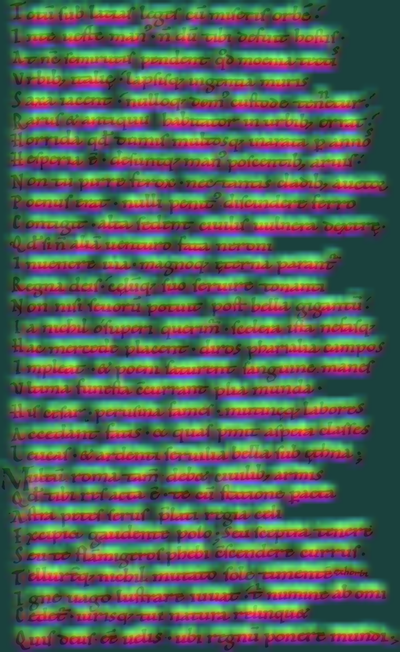} &
\includegraphics[width=0.13\textwidth, height=0.19\textwidth]{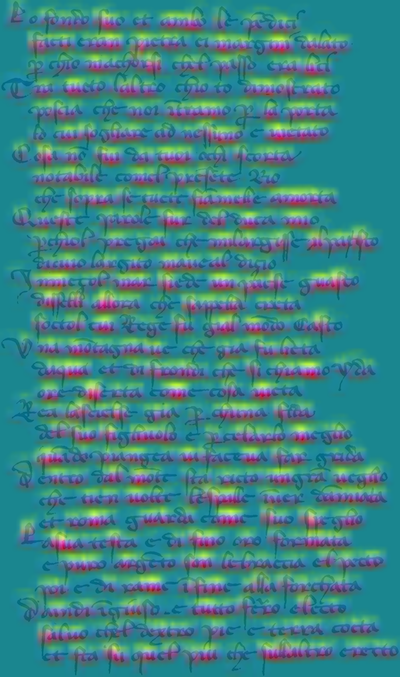} &
\includegraphics[width=0.13\textwidth, height=0.19\textwidth]{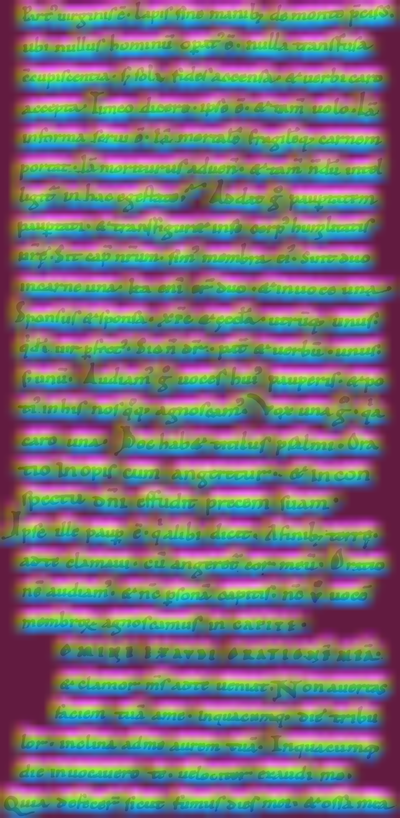}\\
\end{tabular}
\caption{Patch size comparison by qualitative results. Each row shows an example output from different datasets using a patch size. A patch size larger than 400 pixels could not be experimented due to memory overhead. Vertical observation illustrates that the method is insensitive to small variations in the patch size. Very small patches lead blob lines to fade out because they don't contain a coarse trend of text line patterns.}
\label{fig:patch_size}
\end{figure*}

We have experimented with different patch sizes and found $350\times 350$ performs well while keeping memory overhead manageable. \figurename~\ref{fig:patch_size} shows results using patches of variable sizes. One can see that larger patch sizes lead to compact and well separated clusters of blob lines. Obviously at some point the performance is expected to decrease, if the patch size is increased further, because the assumption that the neighbouring patches are similar will gradually decrease. On the other hand the small patches do not contain a coarse trend of text line patterns therefore the blob lines fade out. 

\subsection{Effect of central window size ($w$)}
\begin{figure*}[t]
\centering
\begin{tabular}
{m{0.02\textwidth} m{0.30\textwidth} m{0.30\textwidth} m{0.30\textwidth}}

\multicolumn{1}{c}{} & \multicolumn{1}{c}{20} & \multicolumn{1}{c}{10} & \multicolumn{1}{c}{5} \\

\rotatebox[origin=c]{90}{ICFHR} &
\includegraphics[width=0.30\textwidth, height=0.20\textwidth]{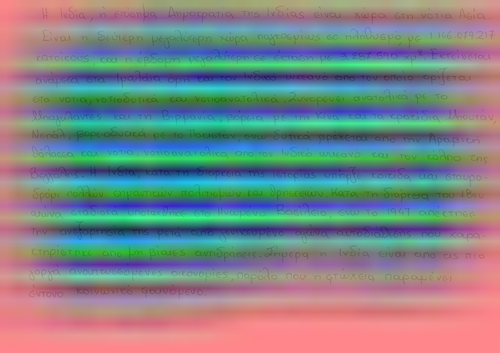} &
\includegraphics[width=0.30\textwidth, height=0.20\textwidth]{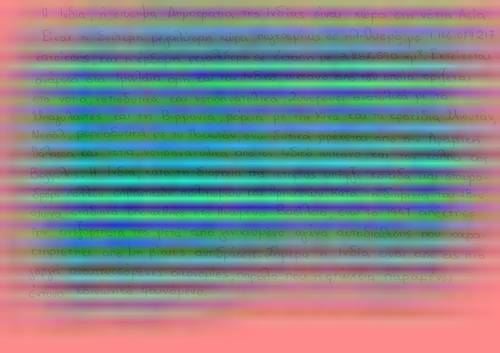} & 
\includegraphics[width=0.30\textwidth, height=0.20\textwidth]{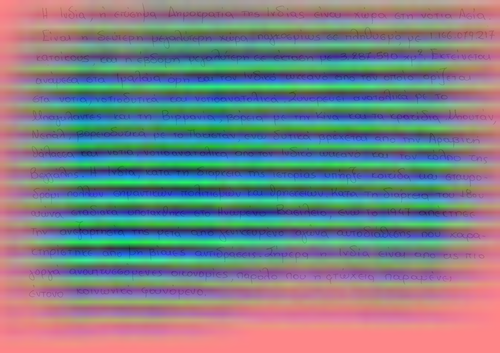} \\

\end{tabular}
\caption{Visualization of the effect of central window size. From left to right shows the results with a decreasing central window size. Central window has to be small enough so the text lines in the downsampled images will not be scrambled. Otherwise blob lines that strike through the text lines will be scrambled.}
\label{fig:central_window}
\end{figure*}

Consider that the input document that is downsampled by a factor of central window size should still be containing the text lines in an apartable form. Input document image size is downsampled by a factor of the central window size of the sliding window. Therefore this factor is effective on the representability of text lines in the pseudo-rgb image. This factor has to be small enough so the text lines in the downsampled images will not be scrambled. Otherwise it is impossible to represent the detected blob lines that strike through the scrambled text lines (\figurename~\ref{fig:central_window}). On the other hand, the computation time is inversely proportional to the central window size. We have experimented with central window sizes and found $w=20$ is efficient and effective well enough.

\subsection{Patch saliency visualization}
\begin{figure*}[t]
\centering
\begin{tabular}
{m{0.02\textwidth} m{0.20\textwidth} m{0.20\textwidth} m{0.20\textwidth}m{0.20\textwidth}}

\multicolumn{1}{c}{} & \multicolumn{1}{c}{Normal} & \multicolumn{1}{c}{Flipped} & \multicolumn{1}{c}{Rotated $180$}& \multicolumn{1}{c}{Rotated $90$} \\

\rotatebox[origin=c]{90}{Pinkas} &
\includegraphics[width=0.20\textwidth, height=0.20\textwidth]{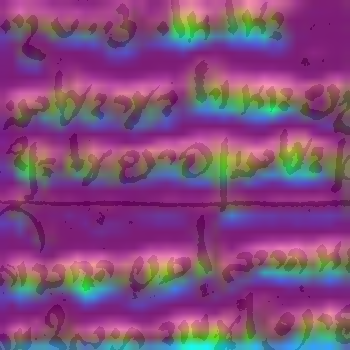} &
\includegraphics[width=0.20\textwidth, height=0.20\textwidth]{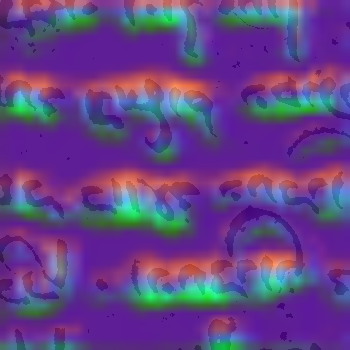} & 
\includegraphics[width=0.20\textwidth, height=0.20\textwidth]{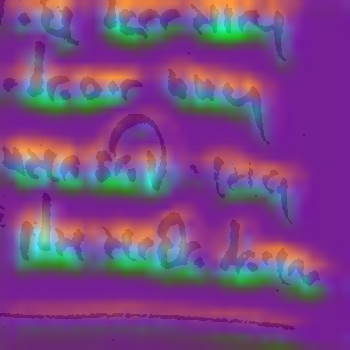} & 
\includegraphics[width=0.20\textwidth, height=0.20\textwidth]{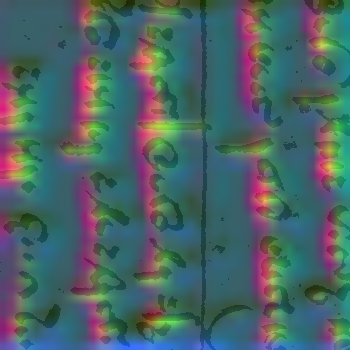} \\

\rotatebox[origin=c]{90}{AHTE} &
\includegraphics[width=0.20\textwidth, height=0.20\textwidth]{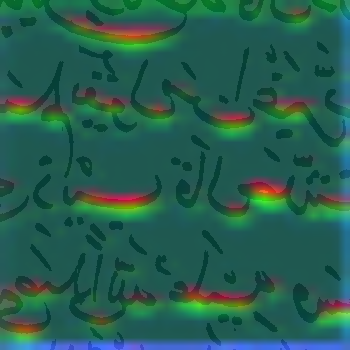} &
\includegraphics[width=0.20\textwidth, height=0.20\textwidth]{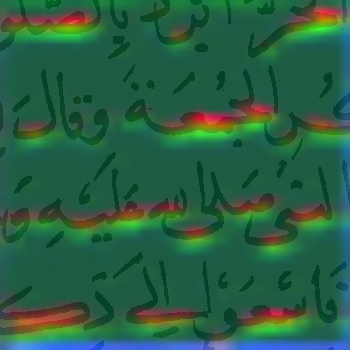} & 
\includegraphics[width=0.20\textwidth, height=0.20\textwidth]{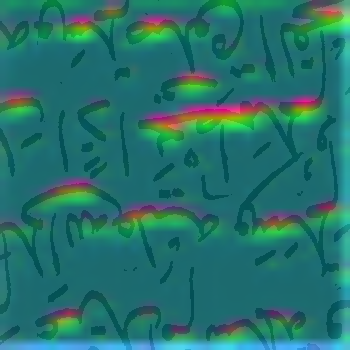} & 
\includegraphics[width=0.20\textwidth, height=0.20\textwidth]{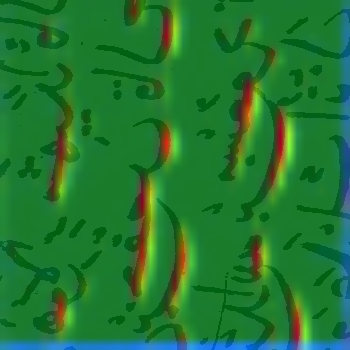} \\

\rotatebox[origin=c]{90}{CSG18} &
\includegraphics[width=0.20\textwidth, height=0.20\textwidth]{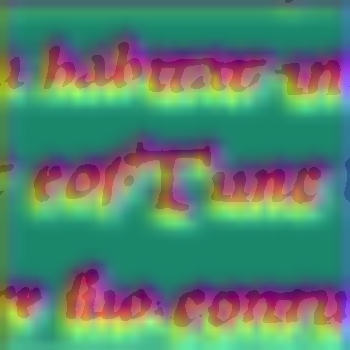} &
\includegraphics[width=0.20\textwidth, height=0.20\textwidth]{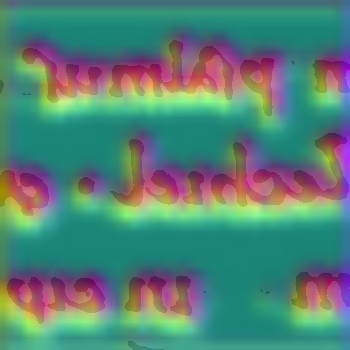} & 
\includegraphics[width=0.20\textwidth, height=0.20\textwidth]{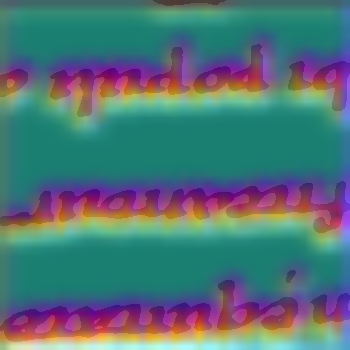} & 
\includegraphics[width=0.20\textwidth, height=0.20\textwidth]{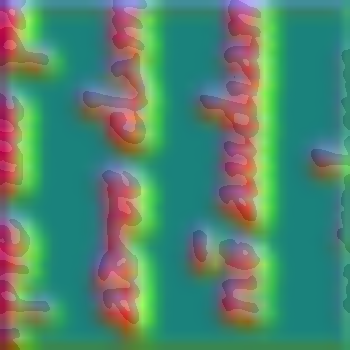} \\

\rotatebox[origin=c]{90}{ICFHR} &
\includegraphics[width=0.20\textwidth, height=0.20\textwidth]{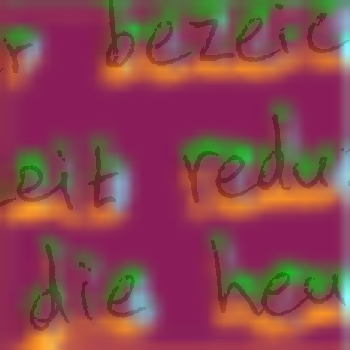} &
\includegraphics[width=0.20\textwidth, height=0.20\textwidth]{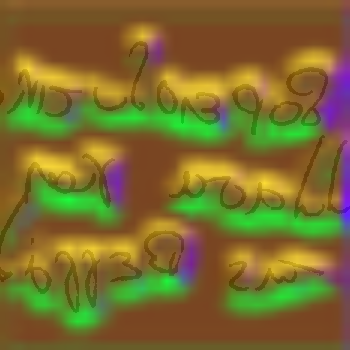} & 
\includegraphics[width=0.20\textwidth, height=0.20\textwidth]{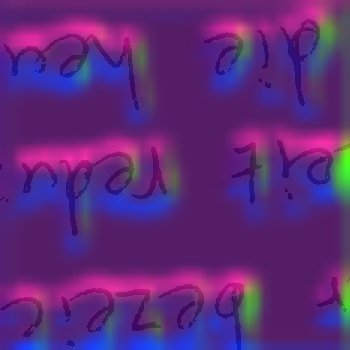} & 
\includegraphics[width=0.20\textwidth, height=0.20\textwidth]{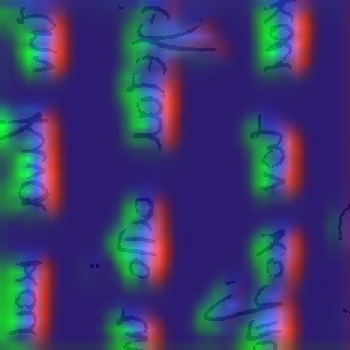} \\

\end{tabular}
\caption{Visualization of the features from the last convolutional layer. No matter the transformation on the patch, the network recognizes the similar salient features on every patch.}
\label{fig:patch_saliency}
\end{figure*}

\begin{figure*}[t]
\centering
\begin{tabular}
{m{0.03\textwidth} m{0.44\textwidth} m{0.44\textwidth} }

\multicolumn{1}{c}{} & \multicolumn{1}{c}{Pinkas} & \multicolumn{1}{c}{AHTE}\\

\rotatebox[origin=c]{90}{Input} &
\includegraphics[width=0.44\textwidth, height=0.31\textwidth]{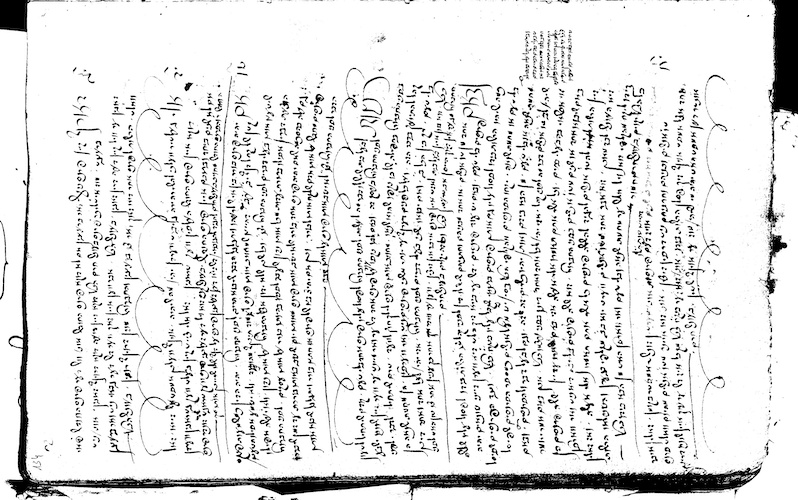} &
\includegraphics[width=0.44\textwidth, height=0.31\textwidth]{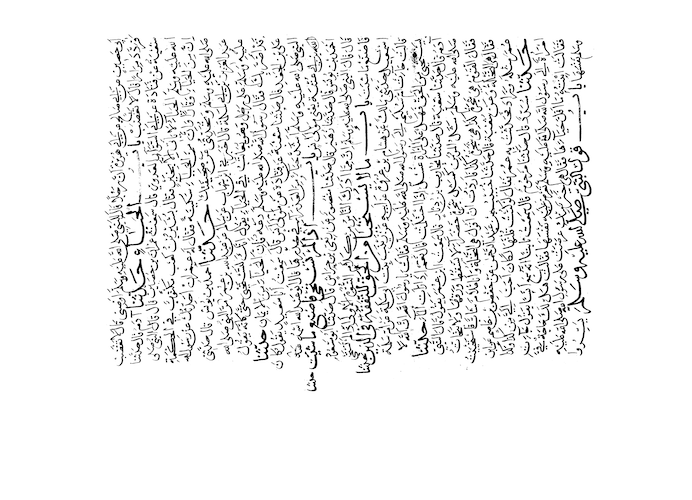}\\

\rotatebox[origin=c]{90}{Output} &
\includegraphics[width=0.44\textwidth, height=0.31\textwidth]{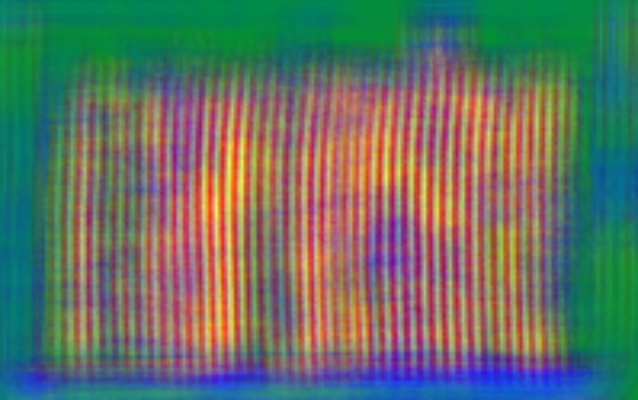} & 
\includegraphics[width=0.44\textwidth, height=0.31\textwidth]{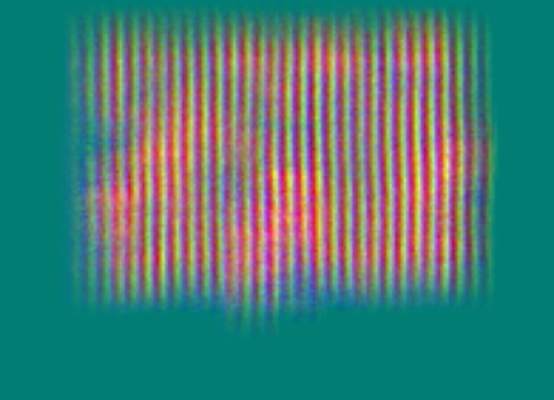}\\

\end{tabular}
\caption{The trained machine can segment an input document image that is entirely rotated by $90$ degrees.}
\label{fig:rotated_input}
\end{figure*}

We visualize the features from last convolutional layer of a single branch to gain insight into the regions that the network looks at the decision of the classifier. The output from the last convolutional layer is a matrix of the size $m \times m \times 512 $ where $m$ is determined by the number of pooling layers and the input patch size $p$. We consider this matrix as $n=m\times m$ vectors each with $512$ dimensions. Then, we get the first three components of these multidimensional vectors and visualize them as a pseudo-rgb image. No matter the transformation on the patch, the network recognizes the similar salient features on every patch (\figurename~\ref{fig:patch_saliency}). As a result of this, it can segment the text lines in a document image that is entirely transformed (\figurename~\ref{fig:rotated_input}).

\subsection{Limitations}
Extracting the features of a document image at patch level is a computationally intensive task and time consuming. Especially the consumed time is inversely proportional to the central window size which has to be small enough to represent the well separated blob lines. Severely skewed or curved text lines do not comply with the assumption that neighbouring patches contain similar coarse trends of text lines. Therefore the method cannot segment a multiply oriented and curved dataset such as the VML-AHTE.

\section{Results}
\label{results}

This section provides quantitative results on the VML-AHTE dataset and the ICDAR 2017 dataset. The results are compared with some other supervised and unsupervised methods. Note that the proposed method uses the same parameters of the baseline experiment on all the datasets. The performance is measured using the text line segmentation evaluation metrics, LIU and PIU, of the ICDAR2017 competition on layout analysis \cite{simistira2017icdar2017}.

\subsection{Results on the VML-AHTE dataset}
We compare our results with those of supervised learning methods, Mask-RCNN \cite{kurar2020text} and FCN+EM \cite{kurar2020text}, and an unsupervised deep learning method, UTLS \cite{kurar2020unsupervised}. Mask-RCNN is an instance segmentation algorithm which is fully supervised using the pixel labels of the text lines. FCN+EM method \cite{kurar2020text} is fully supervised by human annotated blob lines. It uses energy minimization to extract the pixel labels of text lines. The comparison in terms of LIU and PIU are reported in \tablename~\ref{ahte_results}. On the VML-AHTE dataset, the proposed method outperforms the compared methods in terms of LIU metric, and is competitive in terms of the PIU metric. The error cases arise from few number of touching blob lines. Such errors can easily be eliminated but this is out of the focus of this paper. The advantage of the proposed method on the supervised methods is zero labelling effort. Also UTLS \cite{kurar2020unsupervised} has zero labelling effort, however it requires to adjust a heuristic formula. The proposed method eliminates this formula by assuming the neighbouring patches contain the same text line patterns.

\begin{table}[t]
\centering
\caption{LIU and PIU values on the VML-AHTE dataset.}
\label{ahte_results}
\begin{tabular}{@{}rrr@{}}
\toprule
 & LIU & PIU \\ \midrule
\textbf{Unsupervised}\\
UTLS \cite{kurar2020unsupervised}& \textbf{98.55}   &  88.95  \\
Proposed method                  & 90.94            &  83.40  \\
\hline
\textbf{Supervised}\\
Mask-RCNN \cite{kurar2020text}   & 93.08            &  86.97   \\ 
FCN+EM \cite{kurar2020text}      & 94.52            &  \textbf{90.01}  \\
\bottomrule
\end{tabular}
\end{table}

\subsection{Results on the ICDAR2017 dataset}
The second evaluation is carried out on the ICDAR2017 dataset \cite{simistira2017icdar2017}. We run our algorithm on presegmented text block areas by the given ground truth. Hence, we can compare our results with unsupervised System 8 and System 9 which are based on a layout analysis prior to text line segmentation. The comparison in terms of LIU and PIU are reported in \tablename~\ref{icdar_results}. The main challenge in this dataset for the proposed method is the text line parts that are single handed and not accompanied by other text lines in their above and below. Since this is a rare case, the learning system recognizes as an insignificant noise. The performance of the proposed method on the ICDAR dataset is on par with the performances of two unsupervised methods, but these methods probably will need to be readjusted for each new dataset. However, the proposed method has been tested using the same parameters on all the considered datasets.





\newcolumntype{G}{>{\centering\arraybackslash}m{0.22\textwidth}}
\newcolumntype{P}{>{\centering\arraybackslash}m{0.105\textwidth}}
\renewcommand\arraystretch{1.2}
\begin{table*}[t]
\centering
\caption{LIU and PIU values on the ICDAR2017 dataset}
\label{icdar_results}
\begin{tabular}{l P P | P P | P P}
    \toprule
    \multicolumn{1}{l}{} &\multicolumn{2}{G|}{CB55} &  \multicolumn{2}{G|}{CSG18} & \multicolumn{2}{G}{CSG863}\\
    \hline
    \multicolumn{1}{c}{} & LIU & PIU & LIU & PIU & LIU & PIU\\
    \hline
    \textbf{Unsupervised}   &   &   &   &    &   &  \\
    UTLS \cite{kurar2020unsupervised}& 80.35 & 77.30  & 94.30 & 95.50 & 90.58  & 89.40\\
    System-8                & \textbf{99.33} & 93.75 & 94.90 & 94.47  & 96.75 & 90.81\\
    System-9+4.1            & 98.04 & \textbf{96.67}  & 96.91 & \textbf{96.93} & \textbf{98.62} & \textbf{97.54}\\
    Proposed method         & 93.45 & 90.90  & \textbf{97.25} & 96.90  & 92.61  & 91.50\\
     \toprule
\end{tabular}
\end{table*}
\section{Conclusion}
We presented a novel method for unsupervised deep learning of handwritten text line segmentation. It is based on the assumption that in a document image of almost horizontal text lines, the neighbouring patches contain similar coarse pattern of text lines. Hence if one of the neighbouring patches is rotated by $90$ degrees, they contain different coarse pattern of text lines. A network that is trained to embed the similar patches close and the different patches apart in the space, can extract interpretable features for text line segmentation. The method is insensitive to small variations in the input patch size but requires a careful selection of the central window size. We also demonstrated that entirely rotated document images can also be segmented with the same model. The method is effective at detecting cramped, crowded and touching text lines and can surpass the supervised learning methods whereas it has comparable results in terms of text line extraction.
\section*{Acknowledgment}

The authors would like to thank Gunes Cevik and Hamza Barakat for helping in data preparation. This research was partially supported by The Frankel Center for Computer Science at Ben-Gurion University.





%
%
\bibliographystyle{splncs04}
\bibliography{shortbib.bib}
\end{document}